\documentclass[10pt,twocolumn,letterpaper]{article}

\usepackage{cvpr}
\usepackage{times}
\usepackage{epsfig}
\usepackage{graphicx}
\usepackage{amsmath}
\usepackage{amssymb}
\usepackage{bm}

\usepackage{subfigure}
\usepackage{color}
\usepackage{array}
\usepackage{url}
\usepackage{verbatim}
\usepackage{amsthm}
\usepackage{bm}
\usepackage{algorithm}
\usepackage{algorithmicx}
\usepackage{algpseudocode}

\renewcommand{\mathbf}{\boldsymbol}

\usepackage[breaklinks=true,bookmarks=false]{hyperref}

\cvprfinalcopy % *** Uncomment this line for the final submission

\def\cvprPaperID{4680} % *** Enter the CVPR Paper ID here

\ifcvprfinal\fi
\begin{document}

%%%%%%%%% TITLE
\title{Unsupervised Domain Adaptation via Structurally Regularized Deep Clustering}

\author{Hui Tang, Ke Chen, and Kui Jia\thanks{Corresponding author.}\\
South China University of Technology\\
381 Wushan Road, Tianhe District, Guangzhou, Guangdong, China\\
{\tt\small eehuitang@mail.scut.edu.cn}, {\tt\small \{chenk,kuijia\}@scut.edu.cn}
%\and
}

\maketitle
\thispagestyle{empty}

%%%%%%%%% ABSTRACT
\begin{abstract}
Unsupervised domain adaptation (UDA) is to make predictions for unlabeled data on a target domain, given labeled data on a source domain whose distribution shifts from the target one. Mainstream UDA methods learn aligned features between the two domains, such that a classifier trained on the source features can be readily applied to the target ones. However, such a transferring strategy has a potential risk of damaging the \emph{intrinsic} discrimination of target data. To alleviate this risk, we are motivated by the assumption of structural domain similarity, and propose to directly uncover the intrinsic target discrimination via discriminative clustering of target data. We constrain the clustering solutions using structural source regularization that hinges on our assumed structural domain similarity. Technically, we use a flexible framework of deep network based discriminative clustering that minimizes the KL divergence between predictive label distribution of the network and an introduced auxiliary one; replacing the auxiliary distribution with that formed by ground-truth labels of source data implements the structural source regularization via a simple strategy of joint network training. We term our proposed method as \emph{Structurally Regularized Deep Clustering (SRDC)}, where we also enhance target discrimination with clustering of intermediate network features, and enhance structural regularization with soft selection of less divergent source examples. Careful ablation studies show the efficacy of our proposed SRDC. Notably, with no \emph{explicit} domain alignment, SRDC outperforms all existing methods on three UDA benchmarks.

\end{abstract}

%%%%%%%%% BODY TEXT
\section{Introduction}

Given labeled data on a source domain, unsupervised domain adaptation (UDA) is to make predictions in the same label space for unlabeled data on a target domain, where there may exist divergence between the two domains. Mainstream methods are motivated by the classic UDA theories  \cite{da_theory2,da_theory1,mansour09} that specify the learning bounds involving domain divergences, whose magnitudes depend on the feature space and the hypothesis space of classifier. Consequently, these methods (e.g., those recent ones based on adversarial training of deep networks \cite{dann,mcd}) strive to learn aligned features between the two domains, such that classifiers trained on the source features can be readily applied to the target ones. In spite of impressive results achieved by these methods, they have a potential risk of damaging the \emph{intrinsic} structures of target data discrimination, as discussed in \cite{bsp,it_cluster_uda2,da_theory3}. Attempts are made in \cite{bsp,it_cluster_uda2} to alleviate this risk, however, explicit domain alignments are still pursued in their proposed solutions.

To address this issue, we first instantiate the general assumption of domain closeness in UDA problems \cite{da_theory2,it_cluster_uda2} as \emph{structural domain similarity}, which spells as two notions of \emph{domain-wise discrimination} and \emph{class-wise closeness} --- the former notion assumes the existence of intrinsic structures of discriminative data clusters in individual domains, and the later one assumes that clusters of the two domains corresponding to the same class label are geometrically close. This assumption motivates us to consider a UDA approach that directly uncovers the intrinsic data discrimination via discriminative clustering of target data, where we propose to constrain the clustering solutions using structural source regularization hinging on our assumed structural similarity. 

Among various deep network based clustering algorithms \cite{DeepClusterECCVQ18,InfoGAN,DeepClusterRelativeEMICCV17,GMVAE}, we choose a simple but flexible non-generative framework \cite{DeepClusterRelativeEMICCV17}, which performs discriminative clustering by minimizing the KL divergence between predictive label distribution of the network and an introduced auxiliary one. Structural source regularization is simply achieved via a simple strategy of joint network training, by replacing the auxiliary distribution with that formed by ground-truth labels of source data. We term our proposed method as \emph{Structurally Regularized Deep Clustering (SRDC)}. 
In SRDC, we also enhance target discrimination with clustering of intermediate network features, and enhance structural regularization with soft selection of less divergent source examples. We note that quite a few recent UDA methods \cite{cat,can,tpn,dirt_t} consider clustering of target data as well; however, they still do explicit feature alignment between the two domains via alignment of cluster centers/samples, thus prone to the aforementioned risk of damaged intrinsic target discrimination. Experiments on benchmark UDA datasets show the efficacy of our proposed SRDC. We finally summarize our contributions as follows.

\begin{itemize}
\item To address a potential issue of damaging the \emph{intrinsic} data discrimination by explicitly learning domain-aligned features, we propose in this work a source-regularized, deep discriminative clustering method in order to directly uncover the intrinsic discrimination among target data. The method is motivated by our assumption of structural similarity between the two domains, for which we term the proposed method as \emph{Structurally Regularized Deep Clustering (SRDC)}.
    
\item To technically achieve SRDC,  we use a flexible deep clustering framework that first introduces an auxiliary distribution, and then minimizes the KL divergence between the introduced one and the predictive label distribution of the network; replacing the auxiliary distribution with that of ground-truth labels of source data implements the structural source regularization via a simple strategy of joint network training. In SRDC, we also design useful ingredients to enhance target discrimination with clustering of intermediate network features, and to enhance structural regularization with soft selection of less divergent source examples.
    
\item We conduct careful ablation studies on benchmark UDA datasets, which verify the efficacy of individual components proposed in SRDC. Notably, with no \emph{explicit} domain alignment, our proposed SRDC outperforms all existing methods on the benchmark datasets% of Office-31 \cite{office31}, ImageCLEF-DA \cite{imageclefda}, and Office-Home \cite{officehome}
.  
\end{itemize}

\section{Related works}

\noindent\textbf{Alignment based domain adaptation.} A typical line of works \cite{dann, SimNet, adda, iCAN} leverages a domain-adversarial task to align the source and target domains as a whole so that class labels can be transferred from the source domain to the unlabeled target one. Another typical line of works directly minimizes the domain shift measured by various metrics, e.g., maximum mean discrepancy (MMD) \cite{dan, rtn, jan}. These methods are based on domain-level domain alignment. To achieve class-level domain alignment, the works of \cite{cdan,mada} utilize the multiplicative interaction of feature representations and class predictions so that the domain discriminator can be aware of the classification boundary. Based on the integrated task and domain classifier, \cite{dada} encourages a mutually inhibitory relation between category and domain predictions for any input instance. The works of \cite{pfan, cat, tpn, mstn} align the labeled source centroid and pseudo-labeled target centroid of each shared class in the feature space. Some works \cite{swd,adr,mcd} use individual task classifiers for the two domains to detect non-discriminative features and reversely learn a discriminative feature extractor. %The work of \cite{dirt_t} constrains domain-adversarial training by reducing the violation of cluster assumption via entropy minimization. 
Some works \cite{cada,tada, hla} focus attention on transferable regions to derive a domain-invariant classification model. To help achieve target-discriminative features, \cite{deepAdvAttentionAlign, gen_to_adapt} generate synthetic images from the raw input data of the two domains via GANs \cite{gans}. The recent work of \cite{bsp} improves adversarial feature adaptation, where the discriminative structures of target data may be deteriorated \cite{da_theory3}. The work of \cite{larger_norm} adapts the feature norms of the two domains to a large range of values so that the learned features are not only task-discriminative but also domain-invariant.

\noindent\textbf{Clustering based domain adaptation.} The cluster assumption states that the classification boundary should not pass through high-density regions, but instead lie in low-density regions \cite{ClusterAssumption}. To enforce the cluster assumption, conditional entropy minimization \cite{em,min_ent} %, as a typical clustering algorithm \cite{min_ent} or a semi-supervised learning approach \cite{em},
is widely used in the UDA community \cite{rca, it_cluster_uda, dwt_mec, it_cluster_uda2, dirt_t, larger_norm, symnets_v2, symnets}. %By such an information-theoretic criterion, certain predictions for the target data or well-separated target clusters can be achieved.
The work of \cite{can} adopts the spherical $K$-means to assign target labels. The recent work of \cite{cat} employs a Fisher-like criterion based deep clustering loss \cite{sntg}% to encourage the closeness of features from the same class and the separation of features from different classes
. However, they use target clustering just as an incremental technique to improve explicit feature alignment. The previous work of \cite{it_cluster_uda2} is based on the clustering criterion of mutual information maximization, which still explicitly forces domain alignment. In contrast, with no explicit domain alignment, SRDC aims to uncover the intrinsic target discrimination by discriminative target clustering with structural source regularization.

\noindent\textbf{Latent domain discovery.} Methods of latent domain discovery \cite{btda,reshaping_visual_datasets,discover_latent_domains_MSDA,boostDAbyDLD} focus on capturing latent structures of the source, target data or a mixed one under the assumption that data may practically comprise multiple diverse distributions. Our proposed SRDC shares the same motivation with these methods, but differs in the aim to uncover the intrinsic discrimination among target classes by structurally source regularized deep discriminative target clustering, in a distinctive perspective of utilizing structural similarity between the source and target domains.

\section{The strategies of transferring \emph{versus} uncovering the intrinsic target discrimination} % Motivation
\label{SecMotivation}

Consider a source domain $\mathcal{S}$ with $n_s$ labeled examples $\{(\mathbf{x}_j^s, y_j^s)\}_{j=1}^{n_s}$, and a target domain $\mathcal{T}$ with $n_t$ unlabeled examples $\{\mathbf{x}_i^t\}_{i=1}^{n_t}$. Unsupervised domain adaptation (UDA) assumes a shared label space $\mathcal{Y}$ between $\mathcal{S}$ and $\mathcal{T}$. Let $|\mathcal{Y}| = K$ and we have $y^s \in \{1, 2, \dots, K\}$ for any source instance $\mathbf{x}^s$. The objective of \emph{transductive UDA} is to predict $\{\hat{y}_i^t\}_{i=1}^{n_t}$ of $\{\mathbf{x}_i^t\}_{i=1}^{n_t}$ by learning a feature embedding function $\varphi: \mathcal{X} \to \mathcal{Z}$ that lifts any input instance $\mathbf{x} \in \mathcal{X}$ to the feature space $\mathcal{Z}$, and a classifier $f: \mathcal{Z} \to \mathbb{R}^K$. Subtly different from transductive UDA, \emph{inductive UDA} is to measure performance of the learned $\varphi(\cdot)$ and $f(\cdot)$ on held-out instances sampled from the same $\mathcal{T}$. This subtle difference is in fact important since we expect to use the learned $\varphi(\cdot)$ and $f(\cdot)$ as off-the-shelf models, and we expect them to be consistent when learning with different source domains.

\begin{figure}[th]
\begin{center}
\includegraphics[height=0.35\linewidth, width=1.0\linewidth]{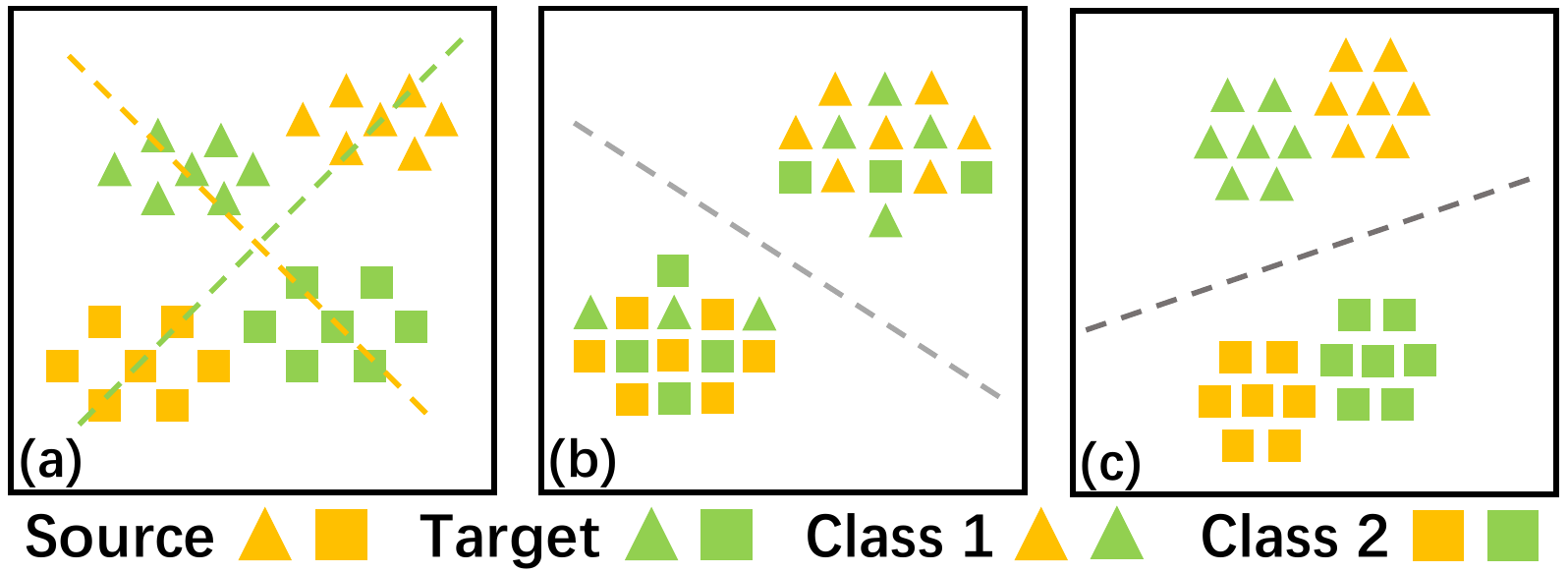}
\end{center} \vspace{-0.3cm}
\caption{(Best viewed in color.) (a) Illustration of the assumption of structural domain similarity (cf. Section \ref{SecMotivation}).  %Data in either the source or target domains form discriminative clusters corresponding to class labels; clusters in the two domains for the same class are spatially adjacent.
The orange line denotes the classifier trained on the labeled source data and the green one denotes the classifier trained on the labeled target data, i.e. the oracle target classifier. 
(b) Illustration of damaging intrinsic structures of data discrimination on the target domain by the existing transferring strategy. The dashed line denotes the source classifier adapting to the damaged discrimination of target data, which has a sub-optimal generalization. (c) Illustration of our proposed uncovering strategy. Discriminative target clustering with structural source regularization uncovers intrinsic target discrimination.}
\label{fig:strategy} \vspace{-0.2cm}
\end{figure}

Domain closeness is generally assumed in UDA either theoretically \cite{da_theory2,mansour09} or intuitively \cite{it_cluster_uda2}. In this work, we summarize the assumptions in \cite{it_cluster_uda2} as the \emph{structural similarity} between the source and target domains, which include the following notions of domain-wise discrimination and class-wise closeness, as illustrated in Figure \ref{fig:strategy}.
\begin{itemize}
\item \emph{Domain-wise discrimination} assumes that there exist intrinsic structures of data discrimination in individual domains, i.e., data in either source or target domains are discriminatively clustered corresponding to the shared label space.

\item \emph{Class-wise closeness} assumes that clusters of the two domains corresponding to the same class label are geometrically close.
\end{itemize}
Based on these assumptions, many of exiting works \cite{dann, cdan, mada, mcd, adda, mdd} take the \emph{transferring} strategy of learning aligned feature representations between the two domains, such that classifiers trained on source features can be readily applied to the target ones. However, such a strategy has a potential risk of damaging the intrinsic data discrimination on the target domain, as discussed in recent works of \cite{bsp,it_cluster_uda2,da_theory3}. An illustration of such damage is also given in Figure \ref{fig:strategy}. We note that more importantly, classifiers adapting to the damaged discrimination of target data would be less effective for tasks of inductive UDA, %since the held-out target data still follow the undamaged, intrinsic data discrimination.
%We note that more importantly, source classifiers adapting to the damaged discrimination of target data would be less effective for tasks of inductive UDA, 
since they deviate too much from the oracle target classifier, i.e. an ideal one trained on the target data with the ground-truth labels.%the held-out target data still follow the undamaged, intrinsic data discrimination.}%{\color{red} 

Based on the above analysis, we are motivated to directly \emph{uncover} the intrinsic target discrimination via discriminative clustering of target data. To leverage the labeled source data, we propose to constrain the clustering solutions using \emph{structural source regularization} that hinges on our assumed structural similarity across domains. Section \ref{SecMethod} presents details of our method, with an illustration given in Figure \ref{fig:strategy}. We note that quite a few recent methods \cite{cat,can,tpn,dirt_t} consider clustering of target data as well; however, they still do explicit feature alignment across domains via alignment of cluster centers/samples, thus prone to the aforementioned risk of damaged intrinsic target discrimination.

\section{Discriminative target clustering with structural source regularization}
\label{SecMethod}

We parameterize the feature embedding function $\varphi(\cdot; \mathbf{\theta})$ and classifier $f(\cdot ; \mathbf{\vartheta})$ as a deep network \cite{resnet,odnn,corr_reg,partnet}, where $\{\mathbf{\theta}, \mathbf{\vartheta}\}$ collects the network parameters. We also write them as  $\varphi(\cdot)$ and $f(\cdot)$ for simplicity, and use $f\circ\varphi$ to denote the whole network. For an input instance $\mathbf{x}$, the network computes feature representation $\mathbf{z} = \varphi(\mathbf{x})$, and outputs a probability vector $\mathbf{p} = \mathrm{softmax}(f(\mathbf{z})) \in [0, 1]^{K}$ after the final softmax operation.

As discussed in Section \ref{SecMotivation}, in order to uncover the intrinsic discrimination of the target domain, we opt for direct clustering of target instances with structural regularization from the source domain. Among various clustering methods \cite{DeepClusterECCVQ18,InfoGAN,DeepClusterRelativeEMICCV17,GMVAE}, we choose a flexible framework of deep discriminative clustering \cite{DeepClusterRelativeEMICCV17}, which minimizes the KL divergence between predictive label distribution of the network and an introduced auxiliary one; by replacing the auxiliary distribution with that of ground-truth labels of source data, we easily implement the structural source regularization via a simple strategy of network joint training, for which we term our proposed method as Structurally Regularized Deep Clustering (SRDC). In SRDC, we also enhance target discrimination with clustering of intermediate network features, and enhance structural regularization with soft selection of less divergent source examples.

\subsection{Deep discriminative target clustering}

For the unlabeled target data $\{\mathbf{x}_i^t\}_{i=1}^{n_t}$, the network predicts, after softmax operation, the probability vectors $\{\mathbf{p}_i^t\}_{i=1}^{n_t}$ that we collectively write as $\mathbf{P}^t$. We also write as $p_{i,k}^t$ the $k^{th}$ element of $\mathbf{p}_i^t$ for the target instance $\mathbf{x}_i^t$. $\mathbf{P}^t$ thus approximates the predictive label distribution of the network for samples of $\mathcal{T}$. Similar to \cite{DeepClusterRelativeEMICCV17,DeepClusterLink}, we first introduce an auxiliary counterpart $\mathbf{Q}^t$, and the proposed SRDC then alternates in (1) updating $\mathbf{Q}^t$, and (2) using the updated $\mathbf{Q}^t$ as labels to train the network to update parameters $\{\mathbf{\theta}, \mathbf{\vartheta}\}$, which optimizes the following objective of deep discriminative clustering
\begin{eqnarray}\label{EqnAuxiTarDistrConstrain}
\begin{aligned}
\min_{\mathbf{Q}^t, \{\mathbf{\theta}, \mathbf{\vartheta}\}} \mathcal{L}_{f\circ\varphi}^t = {\rm KL}(\mathbf{Q}^t||\mathbf{P}^t) + \sum_{k=1}^K {\varrho}_k^t \log {\varrho}_k^t ,
\end{aligned}
\end{eqnarray}
where ${\varrho}_k^t = \frac{1}{n_t} \sum_{i=1}^{n_t} q_{i,k}^t$ and the second term in (\ref{EqnAuxiTarDistrConstrain}) is used to balance cluster assignments in $\{\mathbf{q}_i^t\}_{i=1}^{n_t}$ --- otherwise degenerate solutions would be obtained that merge clusters by removing cluster boundaries \cite{PeronaMIDisCluster}. In addition, it encourages entropy maximization of the label distribution on the target domain, i.e., encouraging cluster {\em size} balance. In aware of the lack of prior knowledge about target label distribution, we simply rely on the second term to account for a uniform one. 
The first term computes the KL divergence between discrete probability distributions $\mathbf{P}^t$ and $\mathbf{Q}^t$ as
\begin{align}
\notag {\rm KL}(\mathbf{Q}^t||\mathbf{P}^t) & = \frac{1}{n_t}\sum_{i=1}^{n_t}\sum_{k=1}^K q_{i,k}^t \log \frac{q_{i,k}^t}{p_{i,k}^t} .
\end{align}
More specifically, the optimization of objective (\ref{EqnAuxiTarDistrConstrain}) takes the following alternating steps.
\begin{itemize}
\item \textbf{Auxiliary distribution update.} Fix network parameters $\{\mathbf{\theta}, \mathbf{\vartheta}\}$ (and $\{\mathbf{p}_i^t\}_{i=1}^{n_t}$ of target instances are fixed as well). By setting the approximate gradient of (\ref{EqnAuxiTarDistrConstrain}) as zero, we has the following closed-form solution \cite{DeepClusterRelativeEMICCV17}
\begin{eqnarray}\label{EqnAuxiTarDistrCloseSol_DisCluster}
\begin{aligned}
q_{i,k}^t = \frac{p_{i,k}^t / (\sum_{i'=1}^{n_t}p_{i',k}^t)^{\frac{1}{2}}}{\sum_{k'=1}^K p_{i,k'}^t / (\sum_{i'=1}^{n_t}p_{i',k'}^t)^{\frac{1}{2}}}.
\end{aligned}
\end{eqnarray}

\item \textbf{Network update.} By fixing $\mathbf{Q}^t$, this step is equivalent to training the network via a cross-entropy loss using $\mathbf{Q}^t$ as labels, giving rise to
\begin{eqnarray}\label{EqnDeepDisTar}
\begin{aligned}
\min\limits_{\mathbf{\theta}, \mathbf{\vartheta}} - \frac{1}{n_t} \sum_{i=1}^{n_t} \sum_{k=1}^{K} q_{i,k}^t \log p_{i,k}^t .
\end{aligned}
\end{eqnarray}
\end{itemize}

In this work, we also enhance uncovering of target discrimination via discriminative clustering in the feature space $\mathcal{Z}$. More specifically, let $\{\mathbf{\mu}_k\}_{k=1}^K$ be the learnable cluster centers of both the source and target data in the space $\mathcal{Z}$. We follow \cite{UnsupervisedEmbeddingICML16} and define a probability vector $\widetilde{\mathbf{p}}_i^t$ of soft cluster assignments of the instance feature $\mathbf{z}_i^t = \varphi(\mathbf{x}_i^t)$ based on instance-to-center distances in the space $\mathcal{Z}$, whose $k^{th}$ element is defined as
\begin{eqnarray}\label{EqnDeepEmbedClusterProb}
\begin{aligned}
\widetilde{p}_{i,k}^t = \frac{\exp((1 + ||\mathbf{z}_i^t - \bm{\mu}_k||^2)^{-1})}{\sum_{k'=1}^K \exp((1 + ||\mathbf{z}_i^t - \bm{\mu}_{k'}||^2)^{-1})} .
\end{aligned}
\end{eqnarray}
We write $\{\widetilde{\mathbf{p}}_i^t\}_{i=1}^{n_t}$ collectively as $\widetilde{\mathbf{P}}^t$. By introducing a corresponding auxiliary distribution $\widetilde{\mathbf{Q}}^t$, we have the following objective of deep discriminative clustering in the space $\mathcal{Z}$
\begin{eqnarray}\label{EqnAuxiTarDistrConstrainFeature}
\begin{aligned}
\min_{\widetilde{\mathbf{Q}}^t, \mathbf{\theta}, \{\mathbf{\mu}_k^t\}_{k=1}^K} \mathcal{L}_{\varphi}^t = {\rm KL}(\widetilde{\mathbf{Q}}^t||\widetilde{\mathbf{P}}^t) + \sum_{k=1}^K \widetilde{\varrho}_k^t \log \widetilde{\varrho}_k^t ,
\end{aligned}
\end{eqnarray}
where $\widetilde{\varrho}_k^t = \frac{1}{n_t} \sum_{i=1}^{n_t} \widetilde{q}_{i,k}^t$. The objective (\ref{EqnAuxiTarDistrConstrainFeature}) can be optimized in the same alternating fashion as for (\ref{EqnAuxiTarDistrConstrain}), by deriving formulations similar to (\ref{EqnAuxiTarDistrCloseSol_DisCluster}) and (\ref{EqnDeepDisTar}), where we note that features $\{\mathbf{z}_i^t\}_{i=1}^{n_t}$ are computed with the updated network parameters $\mathbf{\theta}$, and we also re-initialize $\{\mathbf{\mu}_k\}_{k=1}^K$ at the start of each training epoch based on the current cluster assignments of $\{\mathbf{z}_i^t\}_{i=1}^{n_t}$ (together with labeled source $\{\mathbf{z}_j^s\}_{j=1}^{n_s}$ ). $\{\mathbf{\mu}_k\}_{k=1}^K$ are continuously updated during training iterations of each epoch via back-propagated gradients of (\ref{EqnAuxiTarDistrConstrainFeature}).

Combining (\ref{EqnAuxiTarDistrConstrain}) and (\ref{EqnAuxiTarDistrConstrainFeature}) gives our objective of deep discriminative target clustering, which will be used as the first term of our overall objective of SRDC algorithm
\begin{eqnarray}\label{EqnDeepClusterTar}
\begin{aligned}
\min\limits_{\mathbf{Q}^t, \widetilde{\mathbf{Q}}^t, \{\mathbf{\theta}, \mathbf{\vartheta}\}, \{\mathbf{\mu}_k\}_{k=1}^K} \mathcal{L}_{\textrm{SRDC}}^t = \mathcal{L}_{f\circ\varphi}^t  + \mathcal{L}_{\varphi}^t .
\end{aligned}
\end{eqnarray}

\noindent\textbf{Remarks.} Given unlabeled target data alone, the objective (\ref{EqnAuxiTarDistrConstrain}) itself is not guaranteed to has sensible solutions to uncover the intrinsic discrimination of target data, since the auxiliary distribution $\mathbf{Q}^t$ could be arbitrary whose optimization is subject to no proper constraints. Incorporation of (\ref{EqnAuxiTarDistrConstrainFeature}) into the overall objective (\ref{EqnDeepClusterTar}) would alleviate the issue by soft assignments of $\{\mathbf{z}_i^t\}_{i=1}^{n_t}$ to properly initialized cluster centers $\{\mathbf{\mu}_k\}_{k=1}^K$. To guarantee sensible solutions, deep clustering methods \cite{DeepClusterRelativeEMICCV17,UnsupervisedEmbeddingICML16} usually employ an additional reconstruction loss as a data-dependent regularizer. In our proposed SRDC for domain adaptation, the following introduced structural source regularization serves a similar purpose as that of the reconstruction ones used in \cite{DeepClusterRelativeEMICCV17,UnsupervisedEmbeddingICML16}.

\subsection{Structural source regularization}

Based on the UDA assumption made in Section \ref{SecMotivation} that specifies the structural similarity between the source and target domains, we propose to transfer the global, discriminative structure of labeled source data via a simple strategy of jointly training the same network $f\circ\varphi$. Note that the $K$-way classifier $f$ defines hyperplanes that partition the feature space $\mathcal{Z}$ into regions, of which $K$ ones are uniquely responsible for the $K$ classes. Since the two domains share the same label space, joint training would \emph{ideally} push instances of the two domains from same classes into same regions in $\mathcal{Z}$, thus  \emph{implicitly} achieving feature alignment between the two domains. Figure \ref{fig:strategy} gives an illustration.

Technically, for the labeled source data $\{(\mathbf{x}_j^s, y_j^s)\}_{j=1}^{n_s}$, we simply replace the auxiliary distribution in (\ref{EqnAuxiTarDistrConstrain}) with that formed by the ground-truth labels $\{y_j^s\}_{j=1}^{n_s}$, resulting in a supervised network training via cross-entropy minimization
\begin{eqnarray}\label{EqnDeepDisSrc}
\begin{aligned}
\min\limits_{\mathbf{\theta}, \mathbf{\vartheta}} {\cal{L}}_{f\circ\varphi}^s = - \frac{1}{n_s} \sum_{j=1}^{n_s} \sum_{k=1}^{K} \mathrm{I}[k = y_j^s] \log p_{j,k}^s,
\end{aligned}
\end{eqnarray}
where $p_{j,k}^s$ is the $k^{th}$ element of the predictive probability vector $\mathbf{p}_j^s$ of source instance $\mathbf{x}_j^s$, and $\mathrm{I}[\cdot]$ is the function of indicator. We also enhance source discrimination in the feature space $\mathcal{Z}$, in parallel with (\ref{EqnAuxiTarDistrConstrainFeature}), resulting in
\begin{eqnarray}\label{EqnDeepEmbedSrc}
\begin{aligned}
\min\limits_{\mathbf{\theta}, \{\mathbf{\mu}_k\}_{k=1}^K } {\cal{L}}_{\varphi}^s = - \frac{1}{n_s} \sum_{j=1}^{n_s} \sum_{k=1}^{K} \mathrm{I}[k = y_j^s] \log \widetilde{p}_{j,k}^s ,
\end{aligned}
\end{eqnarray}
where
\begin{align}\label{EqnDeepEmbedClusterProbSrc}
\widetilde{p}_{j,k}^s = \frac{\exp((1 + ||\mathbf{z}_j^s - \bm{\mu}_k ||^2)^{-1})}{\sum_{k'=1}^K \exp((1 + ||\mathbf{z}_j^s - \bm{\mu}_{k'}||^2)^{-1})} .
\end{align}
Combining (\ref{EqnDeepDisSrc}) and (\ref{EqnDeepEmbedSrc}) gives the training objective using labeled source data
\begin{eqnarray}\label{EqnDeepClusterSrc}
\begin{aligned}
\min\limits_{\{\mathbf{\theta}, \mathbf{\vartheta}\}, \{\mathbf{\mu}_k\}_{k=1}^K} \mathcal{L}_{\textrm{SRDC}}^s = \mathcal{L}_{f\circ\varphi}^s  + \mathcal{L}_{\varphi}^s .
\end{aligned}
\end{eqnarray}

Using (\ref{EqnDeepClusterSrc}) as the structural source regularizer, we have our final objective of SRDC algorithm
\begin{eqnarray}\label{EqnDeepJointCluster}
\begin{aligned}
\min\limits_{ \mathbf{Q}^t, \widetilde{\mathbf{Q}}^t, \{\mathbf{\theta}, \mathbf{\vartheta}\}, \{\mathbf{\mu}_k\}_{k=1}^K } {\cal{L}}_{\textrm{SRDC}} = {\cal{L}}_{\textrm{SRDC}}^t + \lambda {\cal{L}}_{\textrm{SRDC}}^s,
\end{aligned}
\end{eqnarray}
where $\lambda$ is a penalty parameter.

\subsection{Enhancement via soft source sample selection}

It is commonly hypothesized in transfer learning \cite{kmm,density_estimate} that importance of source samples varies for learning transferable models. A simple strategy to implement this hypothesis is to re-weight source instances based on their similarities to target ones \cite{pfan,selective_jft,metafgnet}. In this work, we also employ this strategy into SRDC.

Specifically, let $\{\mathbf{c}_k^t \in \mathcal{Z} \}_{k=1}^K$ be the $K$ target cluster centers in the feature space. For any labeled source example $(\mathbf{x}^s, y^s)$, we compute its similarity to $\mathbf{c}_{y^s}^t$, i.e., the target center of cluster $y^s$, based on the following cosine distance
\begin{eqnarray}\label{EqnCosSim}
\begin{aligned}
w^s (\mathbf{x}^s) = \frac{1}{2} \left(1 + \frac{\mathbf{c}_{y^s}^{t\top}\mathbf{x}^s}{||\mathbf{c}_{y^s}^t|| \, ||\mathbf{x}^s||} \right) \in [0, 1] .
\end{aligned}
\end{eqnarray}
We compute $\{\mathbf{c}_k^t \}_{k=1}^K$ once every epoch during network training. Note that $\{\mathbf{c}_k^t \}_{k=1}^K$ are different from $\{\mathbf{\mu}_k \}_{k=1}^K$ in (\ref{EqnDeepEmbedClusterProb}) and (\ref{EqnDeepEmbedClusterProbSrc}), which are cluster centers of both the source and target data that are continuously updated during training iterations of each epoch. We compute weights for all $\{(\mathbf{x}_j^s, y_j^s)\}_{j=1}^{n_s}$ using (\ref{EqnCosSim}), and enhance (\ref{EqnDeepDisSrc}) and (\ref{EqnDeepEmbedSrc}) using the following weighted version of objectives
\begin{eqnarray}\label{EqnDeepDisSrcSoftWeight}
\begin{aligned}
{\cal{L}}_{f\circ\varphi(\cdot; \{w_j^s\}_{j=1}^{n_s})}^s = - \frac{1}{n_s} \sum_{j=1}^{n_s} w_j^s \sum_{k=1}^{K} \mathrm{I}[k = y_j^s] \log p_{j,k}^s,
\end{aligned}
\end{eqnarray}
\begin{eqnarray}\label{EqnDeepEmbedSrcSoftWeight}
\begin{aligned}
{\cal{L}}_{\varphi(\cdot; \{w_j^s\}_{j=1}^{n_s})}^s = - \frac{1}{n_s} \sum_{j=1}^{n_s} w_j^s \sum_{k=1}^{K} \mathrm{I}[k = y_j^s] \log \widetilde{p}_{j,k}^s .
\end{aligned}
\end{eqnarray}
Experiments in Section \ref{SecExp} show that SRDC based on the above weighted objectives achieves improved results.

\section{Experiments}
\label{SecExp}

\subsection{Setups}
\noindent\textbf{Office-31} \cite{office31} is the most popular real-world benchmark dataset for visual domain adaptation, which contains $4,110$ images of $31$ classes shared by three distinct domains: Amazon (\textbf{A}), Webcam (\textbf{W}), and DSLR (\textbf{D}). We evaluate all methods on all the six transfer tasks.

\noindent\textbf{ImageCLEF-DA} \cite{imageclefda} is a benchmark dataset with $12$ classes shared by three domains: Caltech-256 (\textbf{C}), ImageNet ILSVRC 2012 (\textbf{I}), and Pascal VOC 2012 (\textbf{P}). There are $50$ images in each class and $600$ images in each domain. We evaluate all methods on all the six transfer tasks.

\noindent\textbf{Office-Home} \cite{officehome} is a more challenging benchmark dataset, with $15,500$ images of $65$ classes shared by four extremely distinct domains: Artistic images (\textbf{Ar}), Clip Art (\textbf{Cl}), Product images (\textbf{Pr}), and Real-World images (\textbf{Rw}). We evaluate all methods on all the twelve transfer tasks.

\noindent\textbf{Implementation details.} We follow the standard protocol for UDA \cite{dann,tat,cdan,mcd,larger_norm} to use all labeled source samples and all unlabeled target samples as the training data. For each transfer task, we use center-crop target domain images for reporting results and report the classification result of ${\rm mean}$($\pm {\rm std}$) over three random trials. We use the ImageNet \cite{imagenet} pre-trained ResNet-50 \cite{resnet} as the base network, where the last FC layer is replaced with the task-specific FC layer(s) to parameterize the classifier $f(\cdot)$. We implement our experiments in PyTorch. We fine-tune from the pre-trained layers and train the newly added layer(s), where the learning rate of the latter is $10$ times that of the former. We adopt mini-batch SGD with the learning rate schedule as \cite{dann}: the learning rate is adjusted by $\eta_p = \eta_0(1+\alpha p)^{-\beta}$, where $p$ is the process of training epochs normalized to be in $[0,1]$, and $\eta_0=0.001, \alpha=10, \beta=0.75$. We follow \cite{dann} to increase $\lambda$ from $0$ to $1$ by $\lambda_p = 2(1+\exp(-\gamma p))^{-1} - 1$, where $\gamma=10$. The other implementation details are provided in the supplemtary material. The code is available at \url{https://github.com/huitangtang/SRDC-CVPR2020}.

\begin{table*}[th]
\begin{center}
\begin{tabular}{|l|c|c|c|c|c|}%{|p{2.1cm}|p{0.7cm}<{\centering}|p{0.7cm}<{\centering}|p{0.7cm}<{\centering}|p{0.8cm}<{\centering}|p{0.5cm}<{\centering}|}%
\hline
Method                                      & A $\rightarrow$W & A $\rightarrow$D & D $\rightarrow$A & W $\rightarrow$A  & Avg \\
%Method                                     & A $\rightarrow$ & A $\rightarrow$ & D $\rightarrow$ & W $\rightarrow$  & Avg \\
                                            %& W & D & A & A & \\
\hline
\hline
%Source Only                                & 77.8$\pm$0.2 & 82.1$\pm$0.2 & 64.5$\pm$0.2 & 66.1$\pm$0.2 & 72.6 \\
Source Model                                & 77.8$\pm$0.2 & 82.1$\pm$0.2 & 64.5$\pm$0.2 & 66.1$\pm$0.2 & 72.6 \\

%Target Only                                & 85.4$\pm$0.0 & 85.5$\pm$0.0 & 72.2$\pm$0.0 & 73.2$\pm$0.0 & 79.1 \\
%Target only                                & 85.4 & 85.5 & 72.2 & 73.2 & 79.1 \\

%SELECT (w/o source clustering)             & 87.6$\pm$0.1 & 92.2$\pm$0.0 & 73.9$\pm$0.1 & 75.0$\pm$0.1 & 82.2 \\
%DC                             & 87.6 & 92.2 & 73.9 & 75.0 & 82.2 \\
%DC (initialization via source pre-training)  & 87.6$\pm$0.1 & 92.2$\pm$0.0 & 73.9$\pm$0.1 & 75.0$\pm$0.1 & 82.2 \\
SRDC (w/o structural source regularization) & 87.3$\pm$0.0 & 92.1$\pm$0.1 & 73.9$\pm$0.1 & 75.0$\pm$0.1 & 82.1 \\

%SELECT (w/o embedded clustering)           & 94.2$\pm$0.4 & 94.3$\pm$0.4 & 74.3$\pm$0.2 & 75.5$\pm$0.4 & 84.6 \\
SRDC (w/o feature discrimination)           & 94.2$\pm$0.4 & 94.3$\pm$0.4 & 74.3$\pm$0.2 & 75.5$\pm$0.4 & 84.6 \\

%SELECT (w/o source refinement)             & 94.8$\pm$0.2 & 94.6$\pm$0.3 & 74.6$\pm$0.3 & 75.7$\pm$0.3 & 84.9 \\
SRDC (w/o soft source sample selection)            & 94.8$\pm$0.2 & 94.6$\pm$0.3 & 74.6$\pm$0.3 & 75.7$\pm$0.3 & 84.9 \\

%SELECT                                     & \textbf{95.7}$\pm$0.2 & \textbf{95.6}$\pm$0.2 & \textbf{76.5}$\pm$0.3 & \textbf{77.1}$\pm$0.1 & \textbf{86.2} \\
%SRDC                                        & 95.7$\pm$0.2 & 95.6$\pm$0.2 & 76.5$\pm$0.3 & 77.1$\pm$0.1 & 86.2 \\

%SRDC$\rightarrow$AugDC                      & \textbf{96.1}$\pm$0.1 & \textbf{95.8}$\pm$0.1 & \textbf{77.1}$\pm$0.1 & \textbf{77.6}$\pm$0.1 & \textbf{86.7} \\
SRDC                      & \textbf{95.7}$\pm$0.2 & \textbf{95.8}$\pm$0.2 & \textbf{76.7}$\pm$0.3 & \textbf{77.1}$\pm$0.1 & \textbf{86.3} \\

\hline
\end{tabular}
\end{center} \vspace{-0.2cm}
\caption{Ablation studies using Office-31 based on ResNet-50. Please refer to the main text for how different methods are defined.}
\label{table:alter_base_office31} \vspace{-0.3cm}
\end{table*}

\begin{figure}[th]
	\begin{center}
		\includegraphics[height=0.7\linewidth, width=1.0\linewidth]{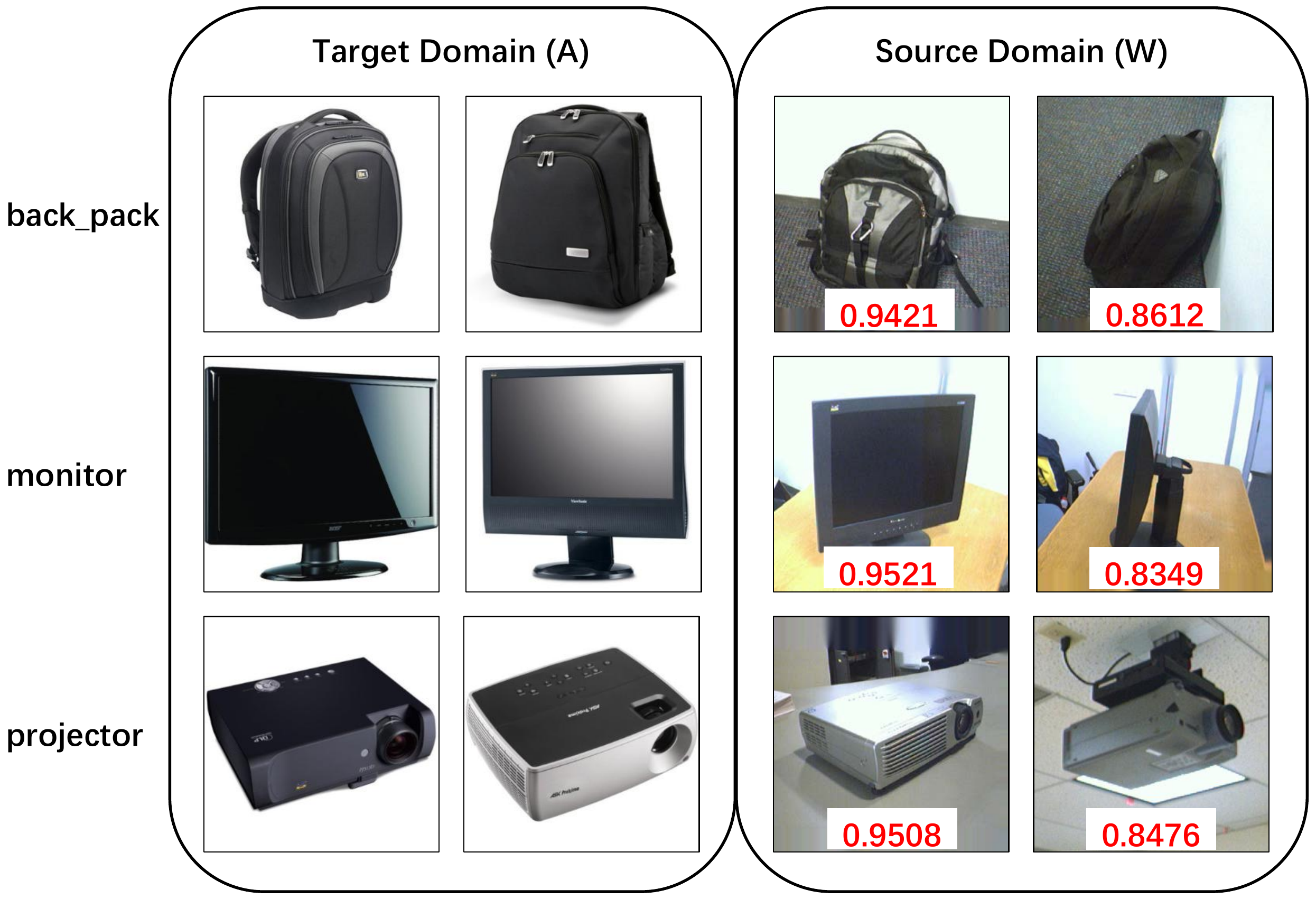}
	\end{center} \vspace{-0.2cm}
	\caption{The images on the left are randomly sampled from the target domain \textbf{A} and those on the right are the top-ranked (the $3^{rd}$ column) and bottom-ranked (the $4^{th}$ column) samples from the source domain \textbf{W} for three classes. Note that the red numbers are the source weights computed by (\ref{EqnCosSim}).}
	\label{fig:select} \vspace{-0.2cm}
\end{figure}

\begin{table}[th]
	\begin{center}
		\resizebox{1.0\linewidth}{!}{
			\begin{tabular}{|l|c|c|c|c|c|}%{|p{2.0cm}|p{0.7cm}<{\centering}|p{0.7cm}<{\centering}|p{0.7cm}<{\centering}|p{0.8cm}<{\centering}|p{0.6cm}<{\centering}|}%
				\hline
				%Method                                      & A $\rightarrow$W & A $\rightarrow$D & D $\rightarrow$A & W $\rightarrow$A  & Avg \\
				Method                                     & A $\rightarrow$W & A $\rightarrow$D & D $\rightarrow$A & W $\rightarrow$A  & Avg \\
				%& W & D & A & A & \\
				\hline
				\hline
				Source Model                                & 79.3 & 81.6 & 63.1 & 65.7 & 72.4 \\
				DANN \cite{dann}                            & 80.8 & 82.4 & 66.0 & 64.6 & 73.5 \\
				MCD \cite{mcd}                              & 86.5 & 86.7 & 72.4 & 70.9 & 79.1 \\
				SRDC                                        & \textbf{91.9} & \textbf{91.6} & \textbf{75.6} & \textbf{75.7} & \textbf{83.7} \\
				\hline
				Oracle Model                                & 98.8 & 97.6 & 87.8 & 87.8 & 93.0 \\
				\hline
			\end{tabular}
		}
	\end{center} \vspace{-0.2cm}
	\caption{Comparative experiments under inductive UDA setting.}
	\label{table:inductive_office31} \vspace{-0.3cm}
\end{table}

\subsection{Ablation studies and analysis}
\noindent\textbf{Ablation study.} To investigate the effects of individual components of our proposed SRDC, we conduct ablation studies using Office-31 based on ResNet-50 by evaluating several variants of SRDC: (1) \textbf{Source Model}, which fine-tunes the base network on labeled source samples; %(2) \textbf{Target only}, which fine-tunes the base network on pseudo-labeled target samples by the initial $K$-means clustering;
(2) \textbf{SRDC (w/o structural source regularization)}%\textbf{DC (initialization via source pre-training)}
, which fine-tunes a source pre-trained model using (\ref{EqnDeepClusterTar}), i.e. without structural source regularization; (3) \textbf{SRDC (w/o feature discrimination)}, which denotes training without source and target discrimination in the feature space $\mathcal{Z}$; (4) \textbf{SRDC (w/o soft source sample selection)}, which denotes training without enhancement via soft source sample selection. The results are reported in Table \ref{table:alter_base_office31}%; (5) \textbf{SRDC}\bm{$\rightarrow$}\textbf{AugDC}, which fine-tunes a pre-trained SRDC model using the proposed deep discriminative target clustering objective (\ref{EqnDeepClusterTar}) with the consistency loss of auxiliary distributions of augmented samples \cite{selfensembling,dwt_mec}
. We can observe that when any one of our designed components is removed, the performance degrades, verifying that (1) both feature discrimination and structural source regularization are effective for improving target clustering; (2) the proposed soft source sample selection scheme leads to better regularization%; (3) reinforcement with source-agnostic target clustering, i.e. SRDC$\rightarrow$AugDC, further improves the clustering solutions
.

\begin{figure*}[th]
\centering
\subfigure[Source Model: \textbf{A}$\to$\textbf{W}]{
\begin{minipage}[t]{0.25\linewidth}
\centering
\includegraphics[width=1.5in]{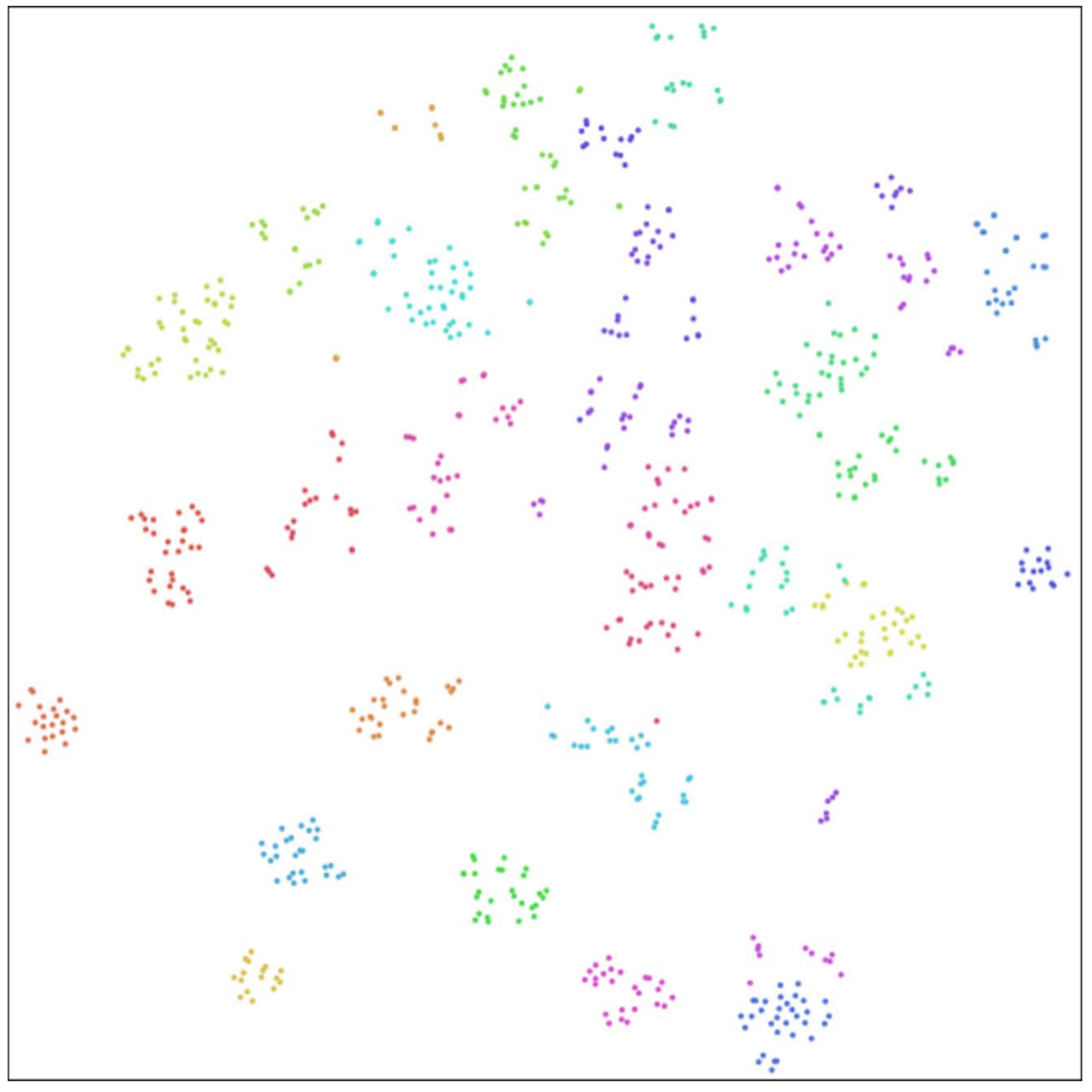}
%\caption{fig1}
\end{minipage}%
}%
\subfigure[SRDC: \textbf{A}$\to$\textbf{W}]{
\begin{minipage}[t]{0.25\linewidth}
\centering
\includegraphics[width=1.5in]{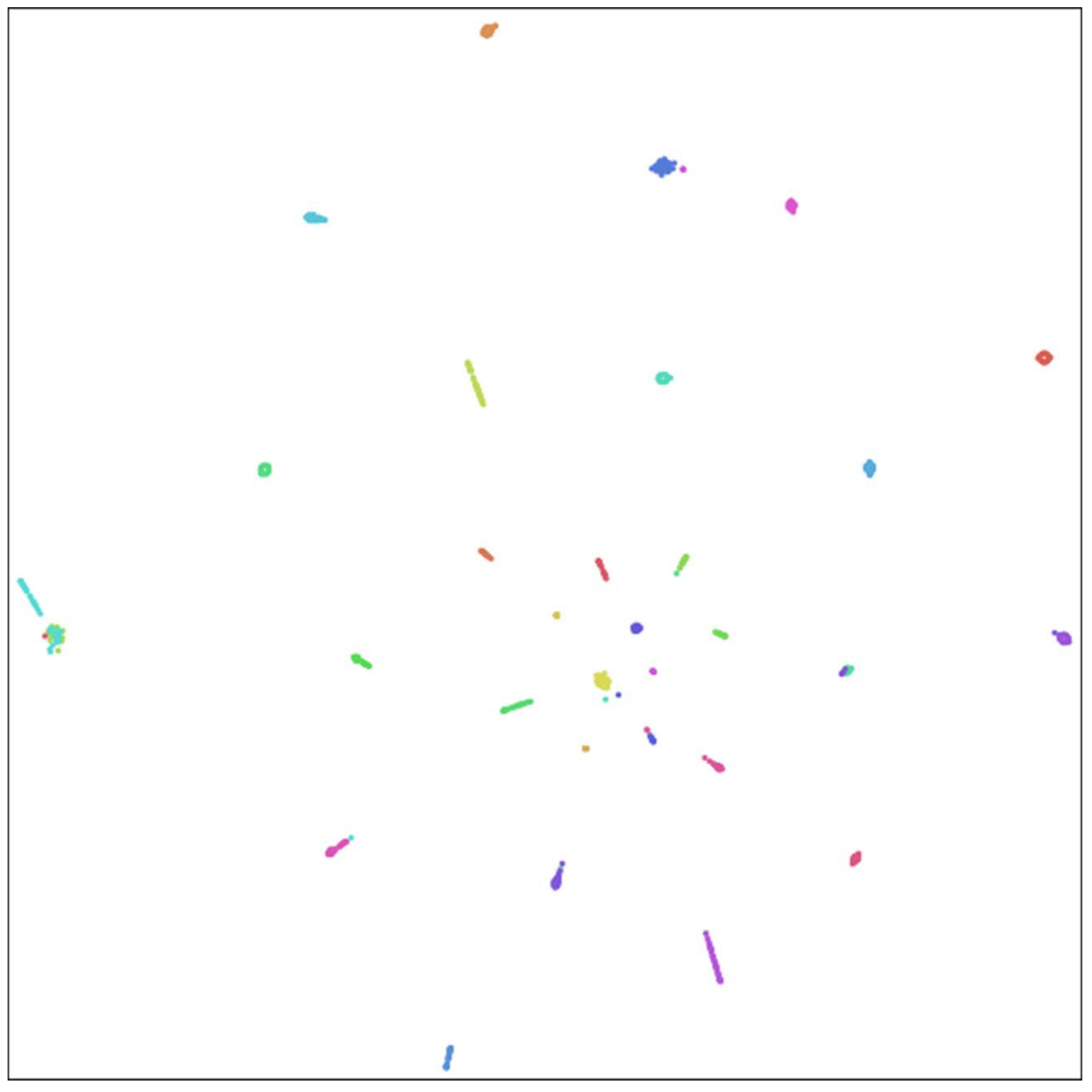}
%\caption{fig2}
\end{minipage}%
}%
\subfigure[Source Model: \textbf{W}$\to$\textbf{A}]{
\begin{minipage}[t]{0.25\linewidth}
\centering
\includegraphics[width=1.5in]{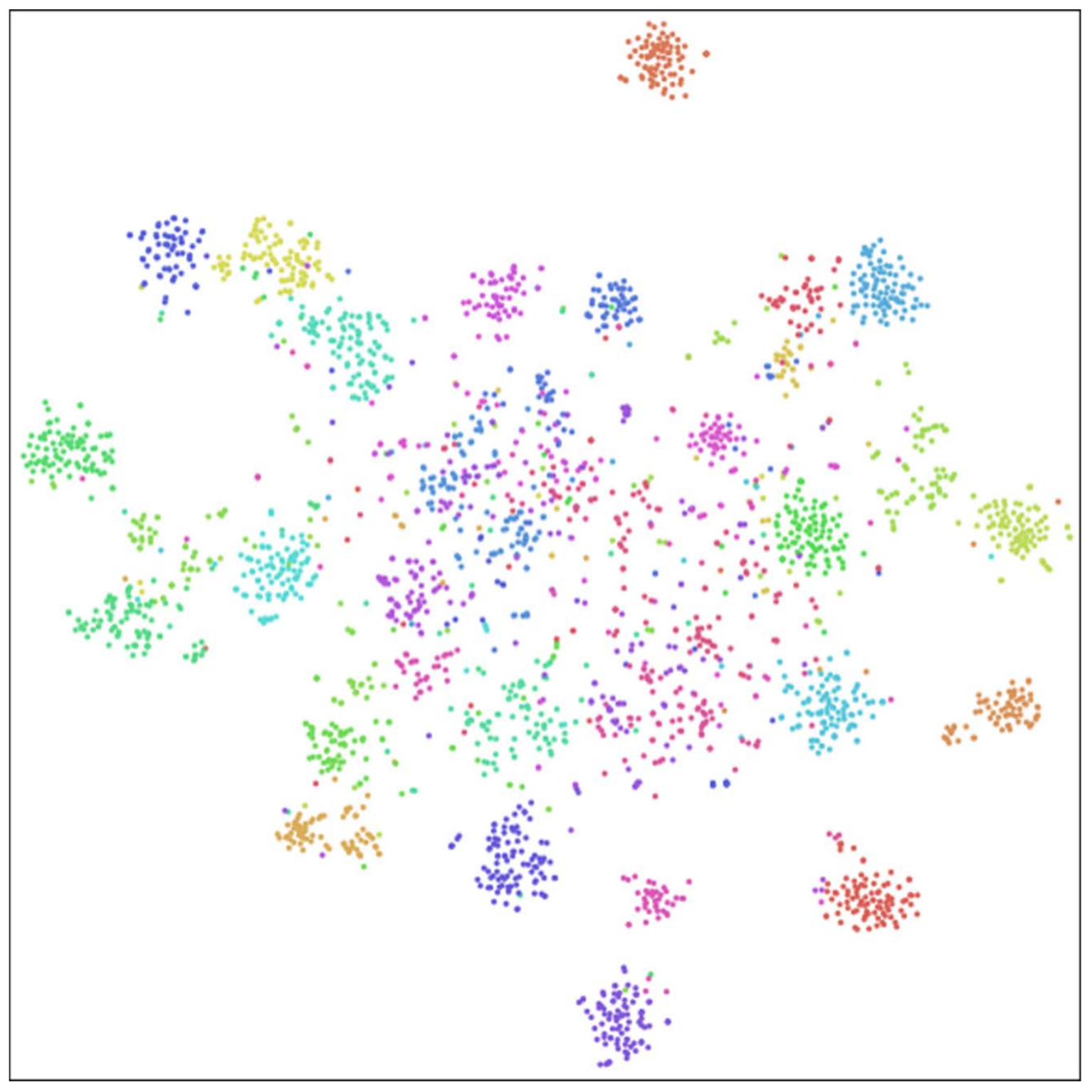}
%\caption{fig2}
\end{minipage}
}%
\subfigure[SRDC: \textbf{W}$\to$\textbf{A}]{
\begin{minipage}[t]{0.25\linewidth}
\centering
\includegraphics[width=1.5in]{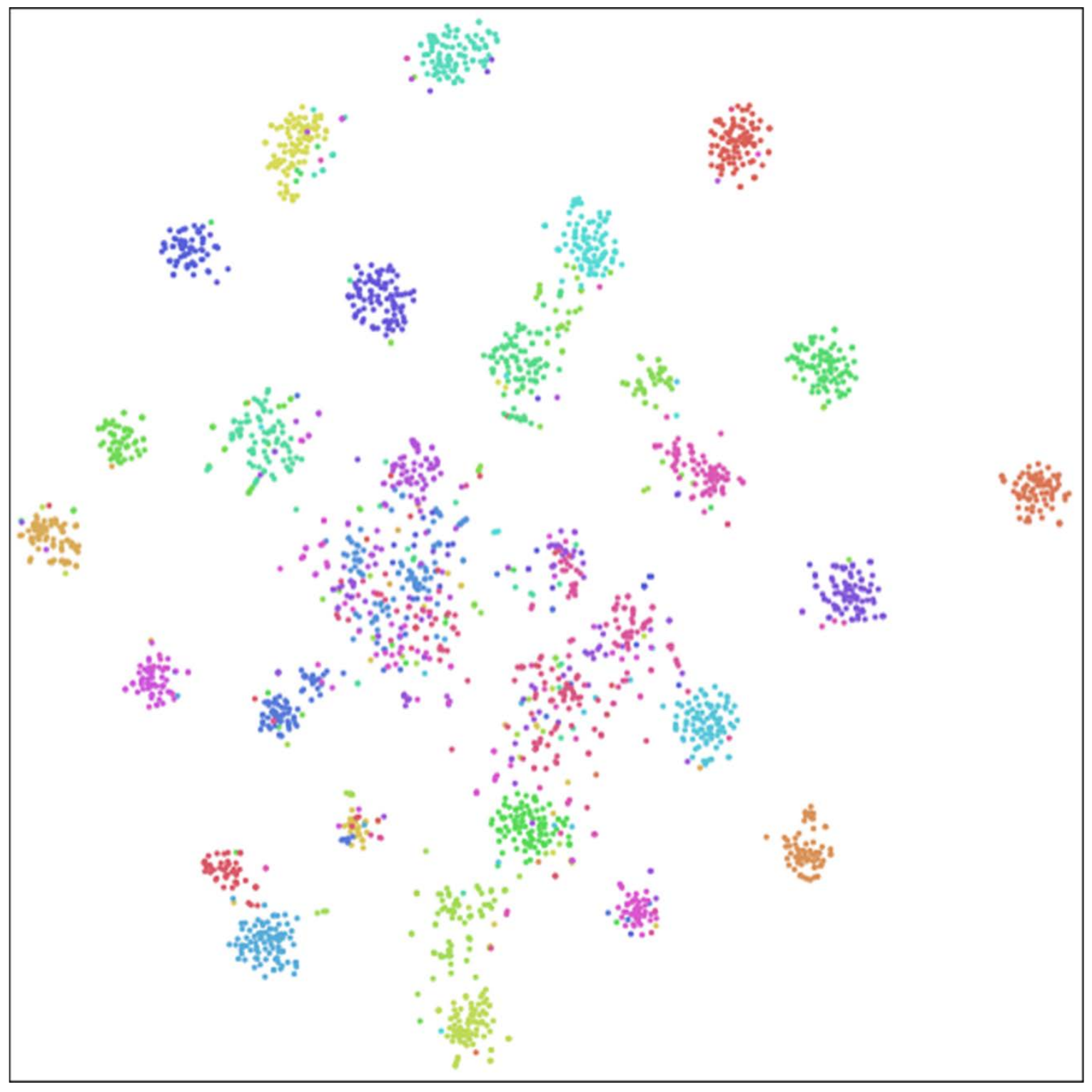}
%\caption{fig2}
\end{minipage}
}%
\centering \vspace{-0.1cm}
\caption{The t-SNE visualization of embedded features on the target domain. Note that different classes are denoted by different colors.}
\label{fig:t_sne} \vspace{-0.4cm}
\end{figure*}

\begin{figure*}[th]
\centering
\subfigure[Source Model: \textbf{A}$\to$\textbf{W}]{
\begin{minipage}[t]{0.25\linewidth}
\centering
\includegraphics[width=1.7in]{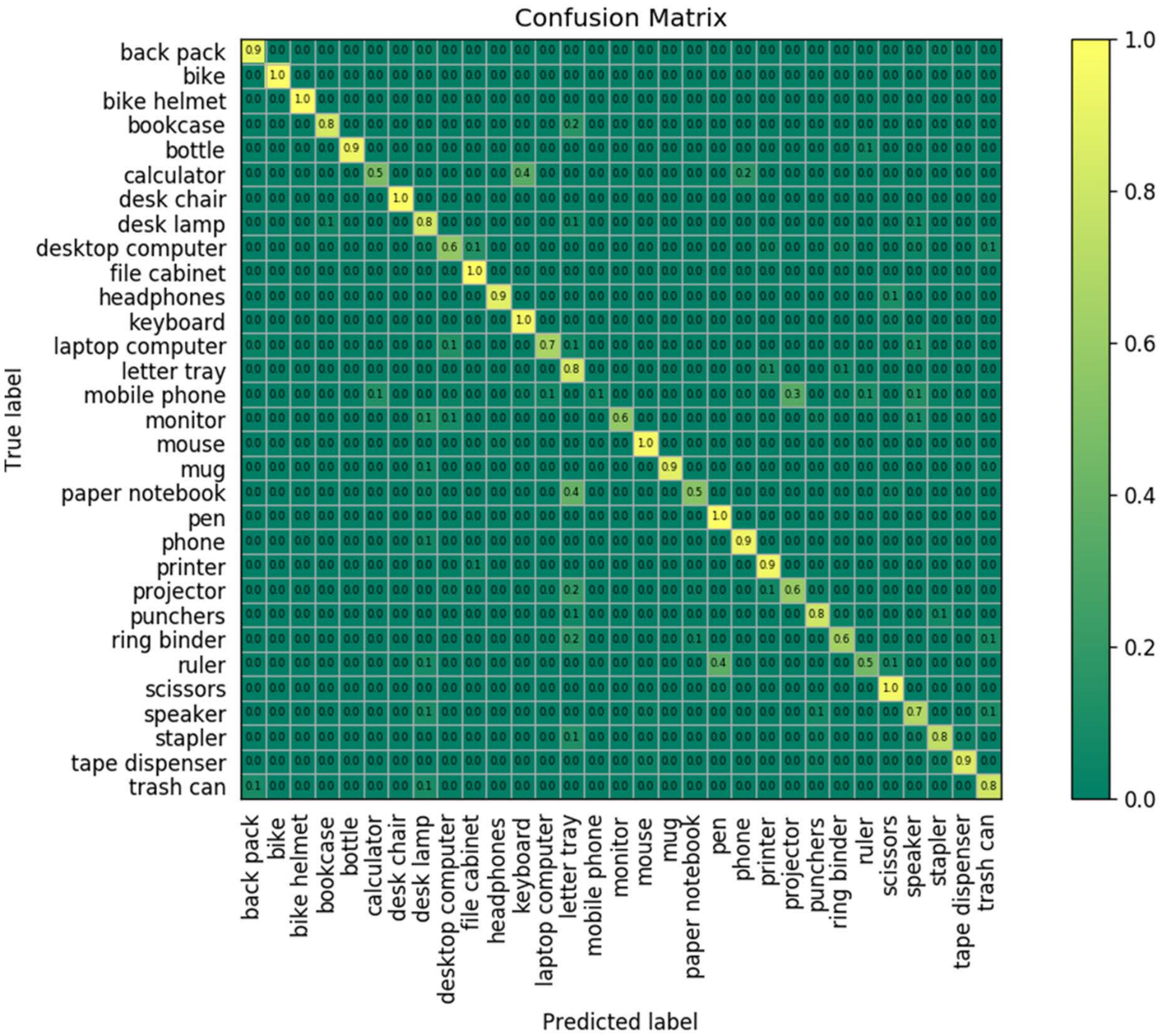}
%\caption{fig1}
\end{minipage}%
}%
\subfigure[SRDC: \textbf{A}$\to$\textbf{W}]{
\begin{minipage}[t]{0.25\linewidth}
\centering
\includegraphics[width=1.7in]{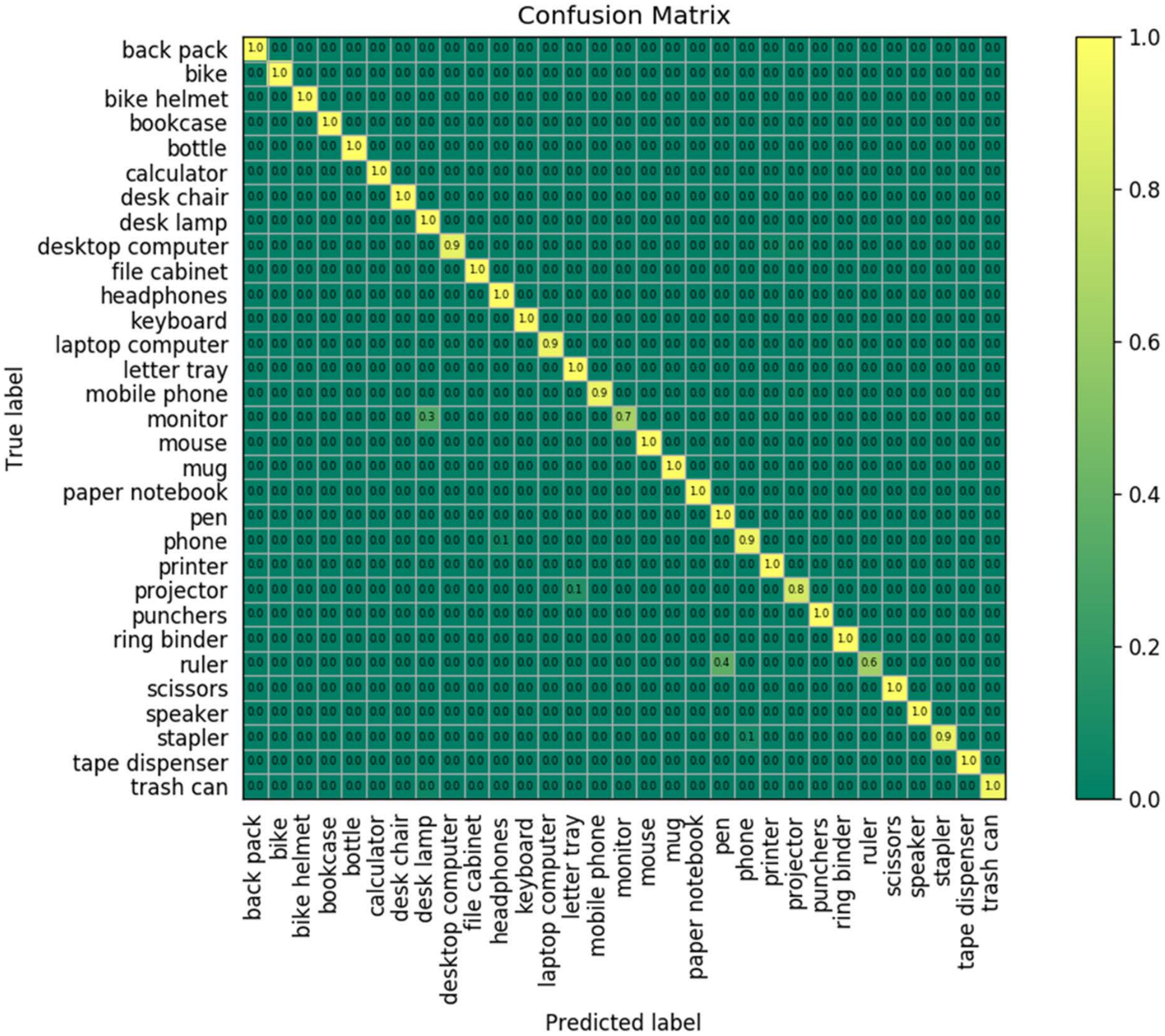}
%\caption{fig2}
\end{minipage}%
}%
\subfigure[Source Model: \textbf{W}$\to$\textbf{A}]{
\begin{minipage}[t]{0.25\linewidth}
\centering
\includegraphics[width=1.7in]{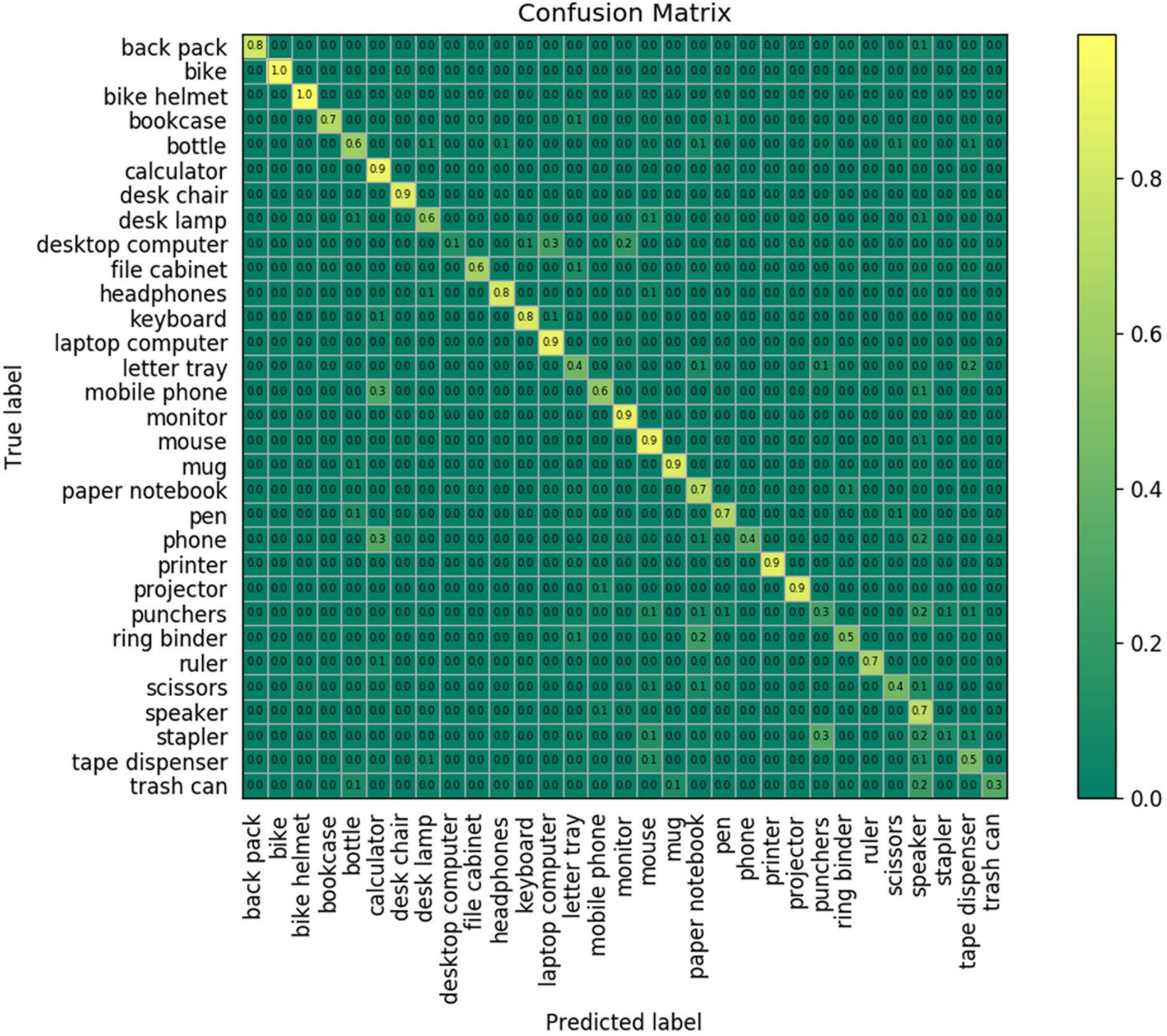}%w2a_sourceonly_cm_v1.png}
%\caption{fig2}
\end{minipage}
}%
\subfigure[SRDC: \textbf{W}$\to$\textbf{A}]{
\begin{minipage}[t]{0.25\linewidth}
\centering
\includegraphics[width=1.7in]{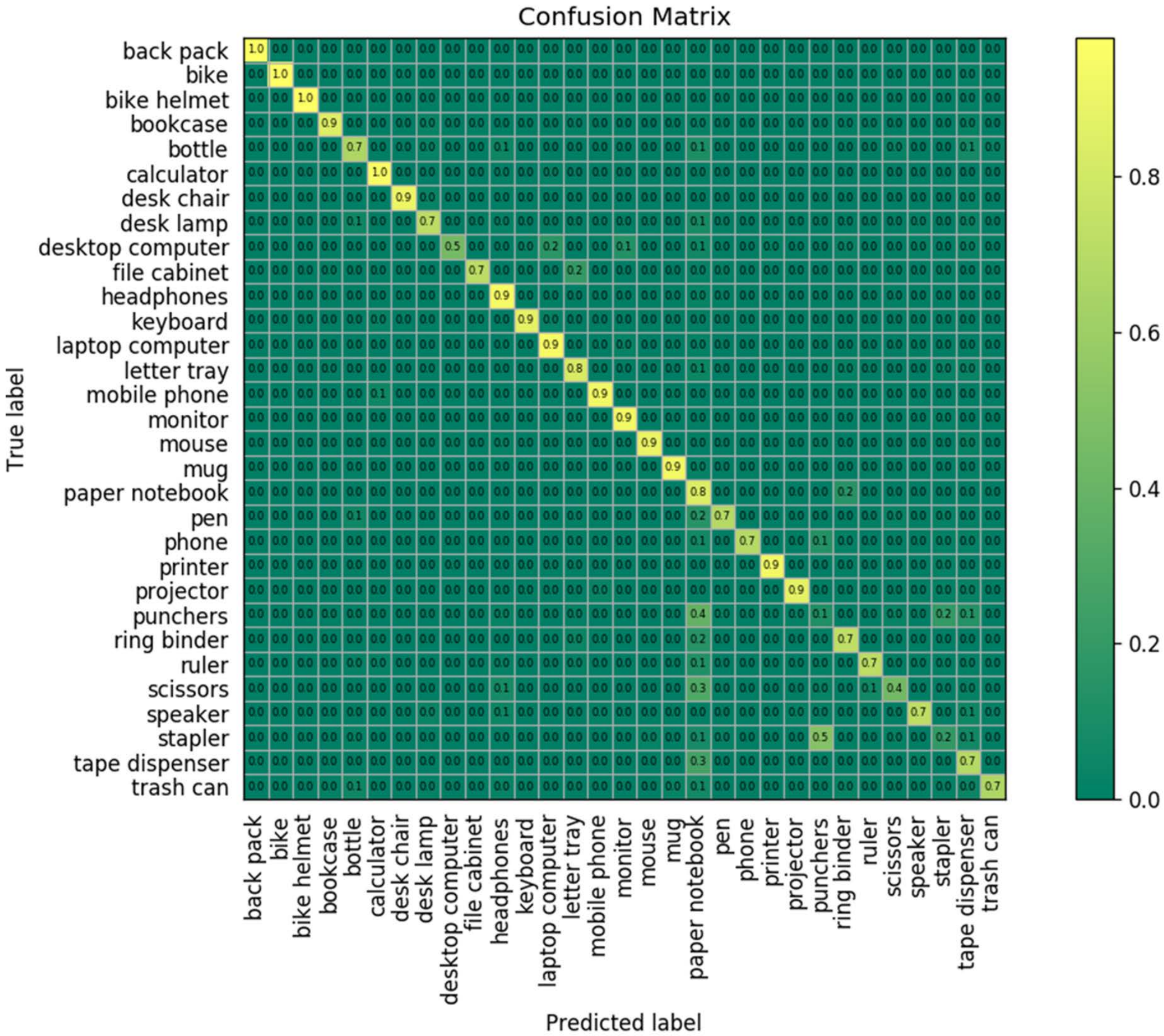}
%\caption{fig2}
\end{minipage}
}%
\centering \vspace{-0.1cm}
\caption{The confusion matrix on the target domain. (Zoom in to see the exact class names!)}
\label{fig:cm} \vspace{-0.3cm}
\end{figure*}

\begin{figure}[ht]
\begin{center}
\includegraphics[height=0.64\linewidth, width=1.0\linewidth]{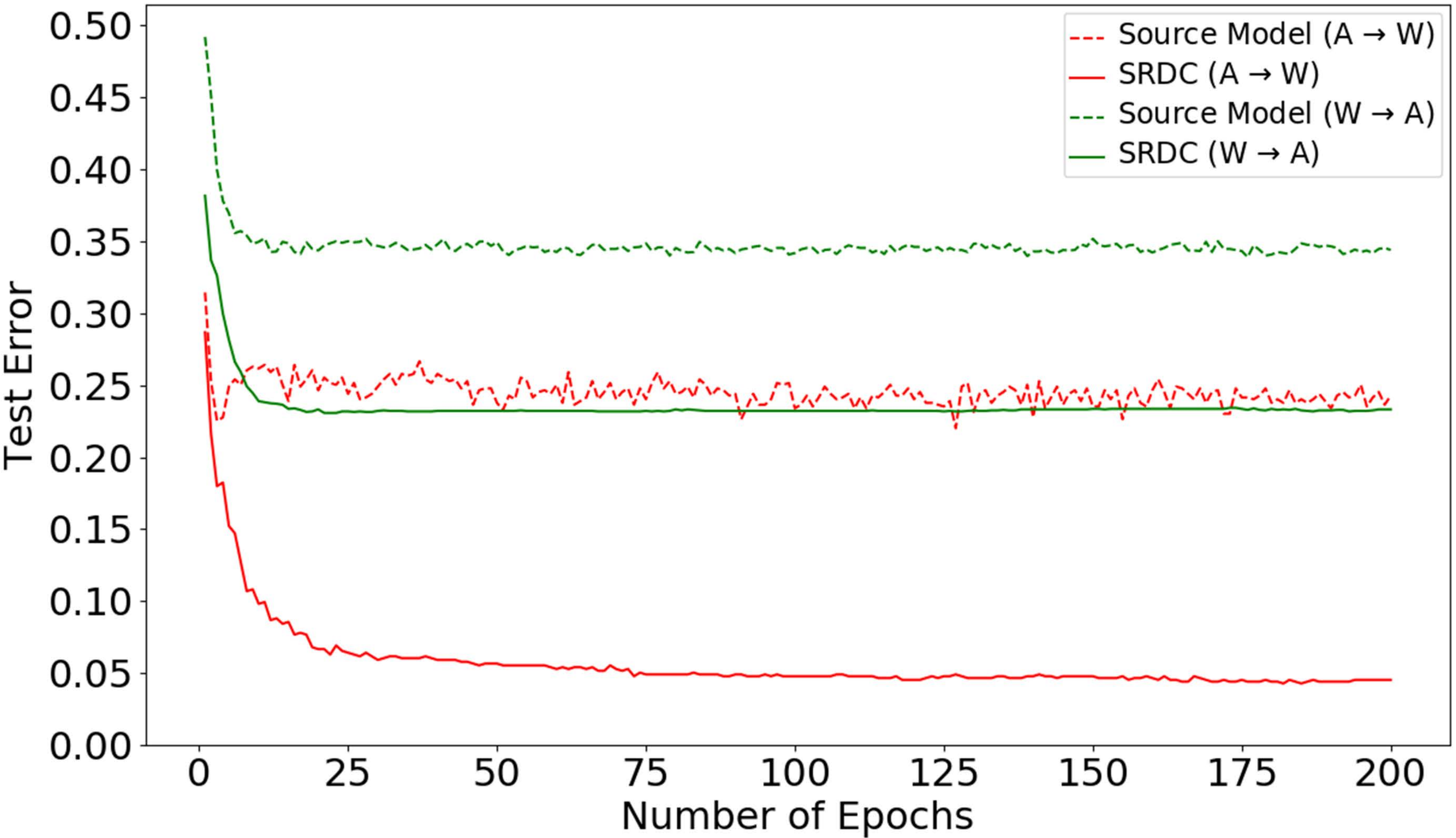}
\end{center} \vspace{-0.4cm}
\caption{Convergence.}
\label{fig:convergence} %\vspace{-0.4cm}
\end{figure}

\noindent\textbf{Source refinement.} To affirm that our proposed soft source sample selection scheme can select more transferable source samples, we show the images randomly sampled from the target domain \textbf{A}, and the top-ranked and bottom-ranked samples from the source domain \textbf{W} in Figure \ref{fig:select}. Here, the red numbers are the source weights computed by (\ref{EqnCosSim}). We can observe that (1) the lowest weight is more than $0.5$, which is reasonable since all source samples are related to the target domain in that the two domains share the same label space; (2) the highest weight is less than $1$, which is reasonable since there exists distribution shift between the two domains; (3) the source images with a canonical viewpoint have the higher weights than those with top-down, bottom-up, and side viewpoints, which is intuitive since all target images are shown only from a canonical viewpoint \cite{office31}. The above observations affirm the rationality of our proposed soft source sample selection scheme.

\noindent\textbf{Comparison under inductive UDA setting.} To verify that our proposed strategy of uncovering the intrinsic target discrimination can derive the clustering solutions closer to the oracle target classifier than the existing transferring strategy of learning
aligned feature representations between the two domains \cite{dann,mcd}, we design comparative experiments under the setting of inductive UDA. We follow a $50\%/50\%$ split scheme to divide each domain of Office-31 into the training and test sets. We use the both labeled sets of the source domain and the unlabeled training set of the target domain as the training data. In Table \ref{table:inductive_office31}, we report results on the test set of the target domain using the best-performing model on the target training set. Here, \textbf{Oracle Model} fine-tunes the base network on the labeled target training set. We can see that our proposed uncovering strategy SRDC achieves closer results to Oracle Model, verifying the motivation of this work and the efficacy of our proposed SRDC.

\noindent\textbf{Feature visualization.} We utilize t-SNE \cite{t_sne} to visualize embedded features on the target domain by Source Model and SRDC for two reverse transfer tasks of \textbf{A}$\to$\textbf{W} and \textbf{W}$\to$\textbf{A} in Figure \ref{fig:t_sne}. We can qualitatively observe that compared to Source Model, the target domain features can be much better discriminated by SRDC, which is based on data clustering to uncover the discriminative data structures.

\noindent\textbf{Confusion matrix.} We give confusion matrixes in terms of accuracy achieved by Source Model and SRDC on two reverse transfer tasks of \textbf{A}$\to$\textbf{W} and \textbf{W}$\to$\textbf{A} in Figure \ref{fig:cm}. Similar to the qualitative result of Figure \ref{fig:t_sne}, we can observe quantitative improvements from Source Model to SRDC, further confirming the advantages of SRDC.

\noindent\textbf{Convergence performance.} We verify the convergence performance of Source Model and SRDC with the test errors on two reverse transfer tasks of \textbf{A}$\to$\textbf{W} and \textbf{W}$\to$\textbf{A} in Figure \ref{fig:convergence}. We can observe that SRDC enjoys faster and smoother convergence performance than Source Model.

%We provide more analysis under synthetic-to-real domain shift and multi-source domain adaptation settings in Section B of the supplementary material. All the empirical evidence confirms the generality and robustness of SRDC.

\begin{table*}[ht]
\begin{center}
	\resizebox{0.8\textwidth}{!}{
\begin{tabular}{|l|c|c|c|c|c|c|c|}
\hline
Method                & A $\rightarrow$ W & D $\rightarrow$ W & W $\rightarrow$ D & A $\rightarrow$ D & D $\rightarrow$ A & W $\rightarrow$ A & Avg \\

\hline
\hline
Source Model \cite{resnet}                  & 77.8$\pm$0.2 & 96.9$\pm$0.1 & 99.3$\pm$0.1 & 82.1$\pm$0.2 & 64.5$\pm$0.2 & 66.1$\pm$0.2 & 81.1 \\

DAN \cite{dan}          & 81.3$\pm$0.3 & 97.2$\pm$0.0 & 99.8$\pm$0.0 & 83.1$\pm$0.2 & 66.3$\pm$0.0 & 66.3$\pm$0.1 & 82.3 \\

DANN \cite{dann}       & 81.7$\pm$0.2 & 98.0$\pm$0.2 & 99.8$\pm$0.0 & 83.9$\pm$0.7 & 66.4$\pm$0.2 & 66.0$\pm$0.3 & 82.6 \\

ADDA \cite{adda}                & 86.2$\pm$0.5 & 96.2$\pm$0.3 & 98.4$\pm$0.3 & 77.8$\pm$0.3 & 69.5$\pm$0.4 & 68.9$\pm$0.5 & 82.9 \\

VADA \cite{dirt_t}              & 86.5$\pm$0.5 & 98.2$\pm$0.4 & 99.7$\pm$0.2 & 86.7$\pm$0.4 & 70.1$\pm$0.4 & 70.5$\pm$0.4 & 85.4 \\

SimNet \cite{SimNet}               & 88.6$\pm$0.5 & 98.2$\pm$0.2 & 99.7$\pm$0.2 & 85.3$\pm$0.3 & 73.4$\pm$0.8 & 71.8$\pm$0.6 & 86.2 \\%$\star$

MSTN \cite{mstn}                   & 91.3 & 98.9 & \textbf{100.0} & 90.4 & 72.7 & 65.6 & 86.5 \\%$\star$ from \cite{dsbn}

GTA \cite{gen_to_adapt} & 89.5$\pm$0.5 & 97.9$\pm$0.3 & 99.8$\pm$0.4 & 87.7$\pm$0.5 & 72.8$\pm$0.3 & 71.4$\pm$0.4 & 86.5 \\

MCD \cite{mcd}                  & 88.6$\pm$0.2 & 98.5$\pm$0.1 & \textbf{100.0}$\pm$0.0 & 92.2$\pm$0.2 & 69.5$\pm$0.1 & 69.7$\pm$0.3 & 86.5 \\

SAFN+ENT \cite{larger_norm}   & 90.1$\pm$0.8 & 98.6$\pm$0.2 & 99.8$\pm$0.0 & 90.7$\pm$0.5 & 73.0$\pm$0.2 & 70.2$\pm$0.3 & 87.1 \\

DAAA \cite{deepAdvAttentionAlign}  & 86.8$\pm$0.2 & \textbf{99.3}$\pm$0.1 & \textbf{100.0}$\pm$0.0 & 88.8$\pm$0.4 & 74.3$\pm$0.2 & 73.9$\pm$0.2 & 87.2 \\ %$\star$

iCAN \cite{iCAN}                & 92.5 & 98.8 & \textbf{100.0} & 90.1 & 72.1 & 69.9 & 87.2 \\

CDAN+E \cite{cdan}  & 94.1$\pm$0.1 & 98.6$\pm$0.1 & \textbf{100.0}$\pm$0.0 & 92.9$\pm$0.2 & 71.0$\pm$0.3 & 69.3$\pm$0.3 & 87.7 \\

MSTN+DSBN \cite{dsbn}              & 92.7 & 99.0 & \textbf{100.0} & 92.2 & 71.7 & 74.4 & 88.3 \\%$\star$

TADA \cite{tada}                & 94.3$\pm$0.3 & 98.7$\pm$0.1 & 99.8$\pm$0.2 & 91.6$\pm$0.3 & 72.9$\pm$0.2 & 73.0$\pm$0.3 & 88.4 \\

TAT \cite{tat}                  & 92.5$\pm$0.3 & \textbf{99.3}$\pm$0.1 & \textbf{100.0}$\pm$0.0 & 93.2$\pm$0.2 & 73.1$\pm$0.3 & 72.1$\pm$0.3 & 88.4 \\

SymNets \cite{symnets}          & 90.8$\pm$0.1 & 98.8$\pm$0.3 & \textbf{100.0}$\pm$0.0 & 93.9$\pm$0.5 & 74.6$\pm$0.6 & 72.5$\pm$0.5 & 88.4 \\ %$\ddagger$

BSP+CDAN \cite{bsp}             & 93.3$\pm$0.2 & 98.2$\pm$0.2 & \textbf{100.0}$\pm$0.0 & 93.0$\pm$0.2 & 73.6$\pm$0.3 & 72.6$\pm$0.3 & 88.5 \\

MDD \cite{mdd}                  & 94.5$\pm$0.3 & 98.4$\pm$0.1 & \textbf{100.0}$\pm$0.0 & 93.5$\pm$0.2 & 74.6$\pm$0.3 & 72.2$\pm$0.1 & 88.9 \\%$\ddagger$

%CADA-P \cite{cada}              & 97.0$\pm$0.2 & 99.3$\pm$0.1 & 100.0$\pm$0.0 & 95.6$\pm$0.1 & 71.5$\pm$0.2 & 73.1$\pm$0.3 & 89.5 \\ %$\star$

CAN \cite{can}                  & 94.5$\pm$0.3 & 99.1$\pm$0.2 & 99.8$\pm$0.2 & 95.0$\pm$0.3 & \textbf{78.0}$\pm$0.3 & 77.0$\pm$0.3 & 90.6 \\ %$\star$
\hline
\textbf{SRDC}                   & \textbf{95.7}$\pm$0.2 & 99.2$\pm$0.1 & \textbf{100.0}$\pm$0.0 & \textbf{95.8}$\pm$0.2 & 76.7$\pm$0.3 & \textbf{77.1}$\pm$0.1 & \textbf{90.8} \\
%\textbf{SRDC}$\bm{\rightarrow}$\textbf{AugDC}    & \textbf{96.1}$\pm$0.1 & 99.2$\pm$0.0 & \textbf{100.0}$\pm$0.0 & \textbf{95.8}$\pm$0.1 & \textbf{77.1}$\pm$0.1 & \textbf{77.6}$\pm$0.1 & \textbf{91.0} \\
%\textbf{SRDC}   & \textbf{96.1}$\pm$0.1 & 99.2$\pm$0.0 & \textbf{100.0}$\pm$0.0 & \textbf{95.8}$\pm$0.1 & \textbf{77.1}$\pm$0.1 & \textbf{77.6}$\pm$0.1 & \textbf{91.0} \\

%\hline

%RTN \cite{rtn}                   & 84.5$\pm$0.2 & 96.8$\pm$0.1 & 99.4$\pm$0.1 & 77.5$\pm$0.3 & 66.2$\pm$0.2 & 64.8$\pm$0.3 & 81.6 \\ %$\dagger$

%JAN-A \cite{jan}                   & 86.0$\pm$0.4 & 96.7$\pm$0.3 & 99.7$\pm$0.1 & 85.1$\pm$0.4 & 69.2$\pm$0.4 & 70.7$\pm$0.5 & 84.6 \\%$\dagger$

%MADA \cite{mada}                   & 90.0$\pm$0.1 & 97.4$\pm$0.1 & 99.6$\pm$0.1 & 87.8$\pm$0.2 & 70.3$\pm$0.3 & 66.4$\pm$0.3 & 85.2 \\%$\dagger$

%\hline

%rRevGrad+CAT \cite{cat}   & 94.4$\pm$0.1 & 98.0$\pm$0.2 & 100.0$\pm$0.0 & 90.8$\pm$1.8 & 72.2$\pm$0.6 & 70.2$\pm$0.1 & 87.6 \\ %tensorflow

\hline
\end{tabular}
}
\end{center} \vspace{-0.3cm}
\caption{Results (\%) on Office-31 (ResNet-50). %Note that all compared methods are based on \textbf{PyTorch} implementation except for the methods in the last six rows based on \textbf{unknown} implementation.
}
\label{table:results_office31} \vspace{-0.2cm}
\end{table*}

\begin{table*}[ht]
\begin{center}
	\resizebox{0.8\textwidth}{!}{
\begin{tabular}{|l|c|c|c|c|c|c|c|}
\hline
Methods                 & I $\rightarrow$ P & P $\rightarrow$ I & I $\rightarrow$ C & C $\rightarrow$ I & C $\rightarrow$ P & P $\rightarrow$ C & Avg \\
\hline
\hline
Source Model \cite{resnet}        & 74.8$\pm$0.3 & 83.9$\pm$0.1 & 91.5$\pm$0.3 & 78.0$\pm$0.2 & 65.5$\pm$0.3 & 91.2$\pm$0.3 & 80.7 \\

DAN \cite{dan}                   & 74.5$\pm$0.4 & 82.2$\pm$0.2 & 92.8$\pm$0.2 & 86.3$\pm$0.4 & 69.2$\pm$0.4 & 89.8$\pm$0.4 & 82.5 \\

DANN \cite{dann}                 & 75.0$\pm$0.6 & 86.0$\pm$0.3 & 96.2$\pm$0.4 & 87.0$\pm$0.5 & 74.3$\pm$0.5 & 91.5$\pm$0.6 & 85.0 \\

JAN \cite{jan}                   & 76.8$\pm$0.4 & 88.0$\pm$0.2 & 94.7$\pm$0.2 & 89.5$\pm$0.3 & 74.2$\pm$0.3 & 91.7$\pm$0.3 & 85.8 \\

%iCAN \cite{iCAN}                 & 79.5 & 89.7 & 94.7 & 89.9 & 78.5 & 92.0 & 87.4 \\

CDAN+E \cite{cdan}               & 77.7$\pm$0.3 & 90.7$\pm$0.2 & 97.7$\pm$0.3 & 91.3$\pm$0.3 & 74.2$\pm$0.2 & 94.3$\pm$0.3 & 87.7 \\

TAT \cite{tat}                   & 78.8$\pm$0.2 & 92.0$\pm$0.2 & 97.5$\pm$0.3 & 92.0$\pm$0.3 & 78.2$\pm$0.4 & 94.7$\pm$0.4 & 88.9 \\

SAFN+ENT \cite{larger_norm}      & 79.3$\pm$0.1 & 93.3$\pm$0.4 & 96.3$\pm$0.4 & 91.7$\pm$0.0 & 77.6$\pm$0.1 & 95.3$\pm$0.1 & 88.9 \\

SymNets \cite{symnets}           & 80.2$\pm$0.3 & 93.6$\pm$0.2 & 97.0$\pm$0.3 & 93.4$\pm$0.3 & 78.7$\pm$0.3 & 96.4$\pm$0.1 & 89.9 \\

\hline

\textbf{SRDC}       & \textbf{80.8}$\pm$0.3 & \textbf{94.7}$\pm$0.2 & \textbf{97.8}$\pm$0.2 & \textbf{94.1}$\pm$0.2 & \textbf{80.0}$\pm$0.3 & \textbf{97.7}$\pm$0.1 & \textbf{90.9} \\

%\textbf{SRDC}       & 80.8$\pm$0.3 & 94.7$\pm$0.2 & 97.8$\pm$0.2 & 94.1$\pm$0.2 & 80.0$\pm$0.3 & 97.7$\pm$0.1 & \textbf{90.9} \\

%\textbf{SRDC}$\bm{\rightarrow}$\textbf{AugDC}       & \textbf{81.0}$\pm$0.1 & \textbf{94.8}$\pm$0.2 & \textbf{97.9}$\pm$0.1 & \textbf{94.2}$\pm$0.1 & \textbf{80.3}$\pm$0.2 & \textbf{98.1}$\pm$0.2 & \textbf{91.1} \\

\hline

%RTN \cite{rtn}                   & 75.6$\pm$0.3 & 86.8$\pm$0.1 & 95.3$\pm$0.1 & 86.9$\pm$0.3 & 72.7$\pm$0.3 & 92.2$\pm$0.4 & 84.9 \\

%MADA \cite{mada}                 & 75.0$\pm$0.3 & 87.9$\pm$0.2 & 96.0$\pm$0.3 & 88.8$\pm$0.3 & 75.2$\pm$0.2 & 92.2$\pm$0.3 & 85.8 \\

%\hline

%rRevGrad+CAT \cite{cat}          & 77.2$\pm$0.2 & 91.0$\pm$0.3 & 95.5$\pm$0.3 & 91.3$\pm$0.3 & 75.3$\pm$0.6 & 93.6$\pm$0.5 & 87.3 \\

%\hline

%CADA-P \cite{cada}               & 78.0 & 90.5 & 96.7 & 92.0 & 77.2 & 95.5 & 88.3 \\

%\hline
\end{tabular}
}
\end{center} \vspace{-0.3cm}
\caption{Results (\%) on ImageCLEF-DA (ResNet-50). %Note that all compared methods are based on \textbf{PyTorch} implementation.
}
\label{table:results_imageclefda} \vspace{-0.2cm}
\end{table*}

\begin{table*}[!ht]
\begin{center}
	\resizebox{1.0\textwidth}{!}{
		\begin{tabular}{|l|c|c|c|c|c|c|c|c|c|c|c|c|c|}%{p{4.5cm}p{0.5cm}<{\centering}p{0.5cm}<{\centering}p{0.5cm}<{\centering}p{0.5cm}<{\centering}p{0.5cm}<{\centering}p{0.5cm}<{\centering}p{0.5cm}<{\centering}p{0.5cm}<{\centering}p{0.5cm}<{\centering}p{0.5cm}<{\centering}p{0.5cm}<{\centering}p{0.5cm}<{\centering}p{0.5cm}<{\centering}}%
		\hline
		Methods                         & Ar$\rightarrow$Cl & Ar$\rightarrow$Pr & Ar$\rightarrow$Rw & Cl$\rightarrow$Ar & Cl$\rightarrow$Pr & Cl$\rightarrow$Rw & Pr$\rightarrow$Ar & Pr$\rightarrow$Cl & Pr$\rightarrow$Rw & Rw$\rightarrow$Ar & Rw$\rightarrow$Cl & Rw$\rightarrow$Pr    & Avg  \\
		                                %& Cl & Pr & Rw & Ar & Pr & Rw & Ar & Cl & Rw & Ar & Cl & Pr &  \\
		\hline
		\hline
		Source Model \cite{resnet}     & 34.9      & 50.0     & 58.0      & 37.4      & 41.9      & 46.2     & 38.5     & 31.2     & 60.4     & 53.9     & 41.2     & 59.9 & 46.1 \\
		
		DAN \cite{dan}                  & 43.6     & 57.0     & 67.9      & 45.8      & 56.5      & 60.4     & 44.0     & 43.6     & 67.7     & 63.1     & 51.5     & 74.3  & 56.3 \\
		
		DANN \cite{dann}     & 45.6     & 59.3     & 70.1      & 47.0      & 58.5      & 60.9     & 46.1     & 43.7     & 68.5     & 63.2     & 51.8      & 76.8 & 57.6 \\
		
		JAN \cite{jan}                  & 45.9     & 61.2     & 68.9      & 50.4      & 59.7      & 61.0     & 45.8     & 43.4     & 70.3     & 63.9     & 52.4      & 76.8 & 58.3 \\
		
		SE \cite{selfensembling}  & 48.8 & 61.8 & 72.8 & 54.1 & 63.2 & 65.1 & 50.6 & 49.2 & 72.3 & 66.1 & 55.9 & 78.7 & 61.5 \\%cited from {dwt_mec}
		
		DWT-MEC \cite{dwt_mec} & 50.3 & 72.1 & 77.0 & 59.6 & 69.3 & 70.2 & 58.3 & 48.1 & 77.3 & 69.3 & 53.6 & 82.0 & 65.6 \\%pytorch
		
		CDAN+E \cite{cdan}  & 50.7     & 70.6     & 76.0     & 57.6       & 70.0      & 70.0     & 57.4     & 50.9     & 77.3      & 70.9      & 56.7     & 81.6 & 65.8 \\
		
		TAT \cite{tat} & 51.6 & 69.5 & 75.4 & 59.4 & 69.5 & 68.6 & 59.5 & 50.5 & 76.8 & 70.9 & 56.6 & 81.6 & 65.8 \\
		
		BSP+CDAN \cite{bsp} & 52.0 & 68.6 & 76.1 & 58.0 & 70.3 & 70.2 & 58.6 & 50.2 & 77.6 & 72.2 & 59.3 & 81.9 & 66.3 \\
		
		SAFN \cite{larger_norm} & 52.0 & 71.7 & 76.3 & 64.2 & 69.9 & 71.9 & 63.7 & 51.4 & 77.1 & 70.9 & 57.1 & 81.5 & 67.3 \\
		
		TADA \cite{tada}       & 53.1 & 72.3 & 77.2 & 59.1 & 71.2 & 72.1 & 59.7 & 53.1 & 78.4 & 72.4 & 60.0 & 82.9 & 67.6 \\
		
		SymNets \cite{symnets} & 47.7 & 72.9 & 78.5 & 64.2 & 71.3 & 74.2 & 64.2 & 48.8 & 79.5 & 74.5 & 52.6 & 82.7 & 67.6 \\
		
		MDD \cite{mdd}         & \textbf{54.9} & 73.7 & 77.8 & 60.0 & 71.4 & 71.8 & 61.2 & 53.6 & 78.1 & 72.5 & \textbf{60.2} & 82.3 & 68.1 \\
		\hline
		\textbf{SRDC} & 52.3 & \textbf{76.3} & \textbf{81.0} & \textbf{69.5} & \textbf{76.2} & \textbf{78.0} & \textbf{68.7} & \textbf{53.8} & \textbf{81.7} & \textbf{76.3} & 57.1 & \textbf{85.0} & \textbf{71.3} \\
		%\textbf{SRDC} & 52.3 & 76.3 & 81.0 & 69.5 & 76.2 & \textbf{78.0} & 68.7 & 53.8 & 81.7 & 76.3 & 57.1 & 85.0 & \textbf{71.3} \\
		%\textbf{SRDC}$\bm{\rightarrow}$\textbf{AugDC} & 52.6 & \textbf{77.5} & \textbf{81.2} & \textbf{69.7} & \textbf{76.5} & \textbf{78.0} & \textbf{68.9} & \textbf{54.0} & \textbf{82.0} & \textbf{76.4} & 57.2 & \textbf{85.2} & \textbf{71.6} \\
		
		\hline
		
		%CADA-P \cite{cada}    & 56.9 & 76.4 & 80.7 & 61.3 & 75.2 & 75.2 & 63.2 & 54.5 & 80.7 & 73.9 & 61.5 & 84.1 & 70.2 \\%$\star$
		
		%\hline
		\end{tabular}
}
\end{center} \vspace{-0.3cm}
\caption{Results (\%) on Office-Home (ResNet-50). %Note that all compared methods are based on \textbf{PyTorch} implementation.
}
\label{table:results_officehome}
\end{table*}

\subsection{Comparisons with the state of the art}
Results on Office-31 based on ResNet-50 are reported in Table \ref{table:results_office31}, where results of existing methods are quoted from their respective papers or the works of \cite{dsbn,tat,cdan}. We can see that SRDC outperforms all compared methods on almost all transfer tasks. It is noteworthy that SRDC significantly enhances the classification results on difficult transfer tasks, e.g. \textbf{A}$\to$\textbf{W} and \textbf{W}$\to$\textbf{A}, where the two domains are quite different. SRDC exceeds the latest work of BSP aiming to improve the discriminability for adversarial feature adaptation, showing that data clustering could be a more promising direction for target discrimination.

Results on ImageCLEF-DA based on ResNet-50 are reported in Table \ref{table:results_imageclefda}, where results of existing methods are quoted from their respective papers or the work of \cite{cdan}. SRDC achieves much better results than all compared methods on all transfer tasks and substantially improves the results on hard transfer tasks, e.g. \textbf{C}$\to$\textbf{P} and \textbf{P}$\to$\textbf{C}, verifying the efficacy of SRDC on transfer tasks with the source and target domains of equal size and class balance.

Results on Office-Home based on ResNet-50 are reported in Table \ref{table:results_officehome}, where results of existing methods are quoted from their respective papers or the works of \cite{cdan, dwt_mec}. We can observe that SRDC significantly exceeds all compared methods on most transfer tasks, with still a large room for improvement. This is reasonable since the four domains in Office-Home contain more categories, are visually more different from each other, and have much lower in-domain classification results \cite{officehome}. It is inspiring that SRDC largely improves over the current state-of-the-art method MDD on such difficult tasks, which underlines the importance of discovering the discriminative structures by data clustering.

%Please refer to Section C in the supplementary material for comparisons with more existing methods on the three benchmark datasets.% and the large-scale dataset of VisDA-2017 \cite{visda2017} with synthetic-to-real domain shift. Please refer to Section C in the supplementary material for comparing SRDC with existing methods under the multi-source domain adaptation setting on the dataset of Office-Caltech10 \cite{gfk}. 

\section{Conclusion}
In this work, motivated by the assumption of structural domain similarity, we propose a source regularized, deep discriminative clustering method, termed as \emph{Structurally Regularized Deep Clustering (SRDC)}. SRDC addresses a potential issue of damaging the intrinsic data discrimination by the existing alignment based UDA methods, via directly uncovering the intrinsic discrimination of target data. Technically, we use a flexible framework of deep network based discriminative clustering that minimizes the KL divergence between predictive label distribution of the network and an introduced auxiliary one; replacing the auxiliary distribution with that formed by ground-truth labels of source data implements the structural source regularization via joint network training. In SRDC, we also enhance target discrimination with clustering of intermediate network features, and enhance structural regularization with soft selection of less divergent source examples. Experiments on benchmarks testify the efficacy of our method.

\noindent\textbf{Acknowledgments.} This work is supported in part by National Natural Science Foundation of China (Grant No.: 61771201), Program for Guangdong Introducing Innovative and Enterpreneurial Teams (Grant No.: 2017ZT07X183), Guangdong R\&D key project of China (Grant No.: 2019B010155001), and Microsoft Research Asia.

{\small
\bibliographystyle{ieee_fullname}
\bibliography{egbib}

\begin{thebibliography}{10}\itemsep=-1pt

\bibitem{imageclefda}
The imageclef-da dataset is available at
  \url{http://imageclef.org/2014/adaptation}.

\bibitem{da_theory2}
Shai Ben-David, John Blitzer, Koby Crammer, Alex Kulesza, Fernando Pereira, and
  Jennifer~Wortman Vaughan.
\newblock A theory of learning from different domains.
\newblock {\em Machine Learning}, 79(1):151--175, May 2010.

\bibitem{da_theory1}
Shai Ben-David, John Blitzer, Koby Crammer, and Fernando Pereira.
\newblock Analysis of representations for domain adaptation.
\newblock In B. Sch\"{o}lkopf, J.~C. Platt, and T. Hoffman, editors, {\em
  Advances in Neural Information Processing Systems 19}, pages 137--144. MIT
  Press, 2007.

\bibitem{DeepClusterECCVQ18}
Mathilde Caron, Piotr Bojanowski, Armand Joulin, and Matthijs Douze.
\newblock Deep clustering for unsupervised learning of visual features.
\newblock In {\em European Conference On Computer Vision}, pages 1692--1700,
  2018.

\bibitem{dsbn}
W. {Chang}, T. {You}, S. {Seo}, S. {Kwak}, and B. {Han}.
\newblock Domain-specific batch normalization for unsupervised domain
  adaptation.
\newblock In {\em 2019 IEEE/CVF Conference on Computer Vision and Pattern
  Recognition (CVPR)}, pages 7346--7354, June 2019.

\bibitem{ClusterAssumption}
O. Chapelle and A. Zien.
\newblock Semi-supervised classification by low density separation.
\newblock In {\em AISTATS 2005}, pages 57--64. Max-Planck-Gesellschaft, Jan.
  2005.

\bibitem{pfan}
C. {Chen}, W. {Xie}, W. {Huang}, Y. {Rong}, X. {Ding}, Y. {Huang}, T. {Xu}, and
  J. {Huang}.
\newblock Progressive feature alignment for unsupervised domain adaptation.
\newblock In {\em 2019 IEEE/CVF Conference on Computer Vision and Pattern
  Recognition (CVPR)}, pages 627--636, June 2019.

\bibitem{InfoGAN}
Xi Chen, Yan Duan, Rein Houthooft, John Schulman, Ilya Sutskever, and Pieter
  Abbeel.
\newblock Infogan: Interpretable representation learning by information
  maximizing generative adversarial nets.
\newblock In D.~D. Lee, M. Sugiyama, U.~V. Luxburg, I. Guyon, and R. Garnett,
  editors, {\em Advances in Neural Information Processing Systems 29}, pages
  2172--2180. Curran Associates, Inc., 2016.

\bibitem{bsp}
Xinyang Chen, Sinan Wang, Mingsheng Long, and Jianmin Wang.
\newblock Transferability vs. discriminability: Batch spectral penalization for
  adversarial domain adaptation.
\newblock In Kamalika Chaudhuri and Ruslan Salakhutdinov, editors, {\em
  Proceedings of the 36th International Conference on Machine Learning},
  volume~97 of {\em Proceedings of Machine Learning Research}, pages
  1081--1090, Long Beach, California, USA, 09--15 Jun 2019. PMLR.

\bibitem{btda}
Z. {Chen}, J. {Zhuang}, X. {Liang}, and L. {Lin}.
\newblock Blending-target domain adaptation by adversarial meta-adaptation
  networks.
\newblock In {\em 2019 IEEE/CVF Conference on Computer Vision and Pattern
  Recognition (CVPR)}, pages 2243--2252, June 2019.

\bibitem{rca}
S. {Cicek} and S. {Soatto}.
\newblock Unsupervised domain adaptation via regularized conditional alignment.
\newblock In {\em 2019 IEEE/CVF International Conference on Computer Vision
  (ICCV)}, pages 1416--1425, Oct 2019.

\bibitem{imagenet}
Jia Deng, Wei Dong, Richard Socher, Li-Jia Li, Kai Li, and Li Fei-Fei.
\newblock Imagenet: A large-scale hierarchical image database.
\newblock In {\em 2009 IEEE Conference on Computer Vision and Pattern
  Recognition}, pages 248--255, 2009.

\bibitem{cat}
Z. {Deng}, Y. {Luo}, and J. {Zhu}.
\newblock Cluster alignment with a teacher for unsupervised domain adaptation.
\newblock In {\em 2019 IEEE/CVF International Conference on Computer Vision
  (ICCV)}, pages 9943--9952, Oct 2019.

\bibitem{DeepClusterRelativeEMICCV17}
K.~G. {Dizaji}, A. {Herandi}, C. {Deng}, W. {Cai}, and H. {Huang}.
\newblock Deep clustering via joint convolutional autoencoder embedding and
  relative entropy minimization.
\newblock In {\em 2017 IEEE International Conference on Computer Vision
  (ICCV)}, pages 5747--5756, Oct 2017.

\bibitem{selfensembling}
Geoff French, Michal Mackiewicz, and Mark Fisher.
\newblock Self-ensembling for visual domain adaptation.
\newblock In {\em International Conference on Learning Representations}, 2018.

\bibitem{dann}
Yaroslav Ganin, Evgeniya Ustinova, Hana Ajakan, Pascal Germain, Hugo
  Larochelle, Fran\c{c}ois Laviolette, Mario Marchand, and Victor Lempitsky.
\newblock Domain-adversarial training of neural networks.
\newblock {\em J. Mach. Learn. Res.}, 17(1):2096--2030, Jan. 2016.

\bibitem{selective_jft}
Weifeng Ge and Yizhou Yu.
\newblock Borrowing treasures from the wealthy: Deep transfer learning through
  selective joint fine-tuning.
\newblock In {\em 2017 IEEE Conference on Computer Vision and Pattern
  Recognition (CVPR)}, pages 10--19, 2017.

\bibitem{reshaping_visual_datasets}
Boqing Gong, Kristen Grauman, and Fei Sha.
\newblock Reshaping visual datasets for domain adaptation.
\newblock In C.~J.~C. Burges, L. Bottou, M. Welling, Z. Ghahramani, and K.~Q.
  Weinberger, editors, {\em Advances in Neural Information Processing Systems
  26}, pages 1286--1294. Curran Associates, Inc., 2013.

\bibitem{gans}
Ian Goodfellow, Jean Pouget-Abadie, Mehdi Mirza, Bing Xu, David Warde-Farley,
  Sherjil Ozair, Aaron Courville, and Yoshua Bengio.
\newblock Generative adversarial nets.
\newblock In {\em Advances in Neural Information Processing Systems 27}, pages
  2672--2680. 2014.

\bibitem{em}
Yves Grandvalet and Yoshua Bengio.
\newblock Semi-supervised learning by entropy minimization.
\newblock In {\em Proceedings of the 17th International Conference on Neural
  Information Processing Systems}, NIPS'04, pages 529--536, Cambridge, MA, USA,
  2004. MIT Press.

\bibitem{resnet}
K. {He}, X. {Zhang}, S. {Ren}, and J. {Sun}.
\newblock Deep residual learning for image recognition.
\newblock In {\em 2016 IEEE Conference on Computer Vision and Pattern
  Recognition (CVPR)}, pages 770--778, June 2016.

\bibitem{discover_latent_domains_MSDA}
Judy Hoffman, Brian Kulis, Trevor Darrell, and Kate Saenko.
\newblock Discovering latent domains for multisource domain adaptation.
\newblock In Andrew Fitzgibbon, Svetlana Lazebnik, Pietro Perona, Yoichi Sato,
  and Cordelia Schmid, editors, {\em Computer Vision -- ECCV 2012}, pages
  702--715, Berlin, Heidelberg, 2012. Springer Berlin Heidelberg.

\bibitem{kmm}
Jiayuan Huang, Arthur Gretton, Karsten Borgwardt, Bernhard Sch\"{o}lkopf, and
  Alex~J. Smola.
\newblock Correcting sample selection bias by unlabeled data.
\newblock In B. Sch\"{o}lkopf, J.~C. Platt, and T. Hoffman, editors, {\em
  Advances in Neural Information Processing Systems 19}, pages 601--608. MIT
  Press, 2007.

\bibitem{DeepClusterLink}
Mohammed Jabi, Marco Pedersoli, Amar Mitiche, and Ismail Ben~Ayed.
\newblock Deep clustering: On the link between discriminative models and
  k-means.
\newblock {\em CoRR}, arXiv:1810.04246, 2018.

\bibitem{odnn}
K. {Jia}, S. {Li}, Y. {Wen}, T. {Liu}, and D. {Tao}.
\newblock Orthogonal deep neural networks.
\newblock {\em IEEE Transactions on Pattern Analysis and Machine Intelligence},
  pages 1--1, 2019.

\bibitem{corr_reg}
K. {Jia}, J. {Lin}, M. {Tan}, and D. {Tao}.
\newblock Deep multi-view learning using neuron-wise correlation-maximizing
  regularizers.
\newblock {\em IEEE Transactions on Image Processing}, 28(10):5121--5134, Oct
  2019.

\bibitem{can}
G. {Kang}, L. {Jiang}, Y. {Yang}, and A.~G. {Hauptmann}.
\newblock Contrastive adaptation network for unsupervised domain adaptation.
\newblock In {\em 2019 IEEE/CVF Conference on Computer Vision and Pattern
  Recognition (CVPR)}, pages 4888--4897, June 2019.

\bibitem{deepAdvAttentionAlign}
Guoliang Kang, Liang Zheng, Yan Yan, and Yi Yang.
\newblock Deep adversarial attention alignment for unsupervised domain
  adaptation: The benefit of target expectation maximization.
\newblock In Vittorio Ferrari, Martial Hebert, Cristian Sminchisescu, and Yair
  Weiss, editors, {\em Computer Vision -- ECCV 2018}, pages 420--436, Cham,
  2018. Springer International Publishing.

\bibitem{PeronaMIDisCluster}
Andreas Krause, Pietro Perona, and Ryan~G. Gomes.
\newblock Discriminative clustering by regularized information maximization.
\newblock In J.~D. Lafferty, C.~K.~I. Williams, J. Shawe-Taylor, R.~S. Zemel,
  and A. Culotta, editors, {\em Advances in Neural Information Processing
  Systems 23}, pages 775--783. 2010.

\bibitem{cada}
V.~K. {Kurmi}, S. {Kumar}, and V.~P. {Namboodiri}.
\newblock Attending to discriminative certainty for domain adaptation.
\newblock In {\em 2019 IEEE/CVF Conference on Computer Vision and Pattern
  Recognition (CVPR)}, pages 491--500, June 2019.

\bibitem{swd}
C. {Lee}, T. {Batra}, M.~H. {Baig}, and D. {Ulbricht}.
\newblock Sliced wasserstein discrepancy for unsupervised domain adaptation.
\newblock In {\em 2019 IEEE/CVF Conference on Computer Vision and Pattern
  Recognition (CVPR)}, pages 10277--10287, June 2019.

\bibitem{min_ent}
Haifeng Li, Keshu Zhang, and Tao Jiang.
\newblock Minimum entropy clustering and applications to gene expression
  analysis.
\newblock In {\em Proceedings of 2004 IEEE Computational Systems Bioinformatics
  Conference}, pages 142--151, 2004.

\bibitem{tat}
Hong Liu, Mingsheng Long, Jianmin Wang, and Michael Jordan.
\newblock Transferable adversarial training: A general approach to adapting
  deep classifiers.
\newblock In Kamalika Chaudhuri and Ruslan Salakhutdinov, editors, {\em
  Proceedings of the 36th International Conference on Machine Learning},
  volume~97 of {\em Proceedings of Machine Learning Research}, pages
  4013--4022, Long Beach, California, USA, 09--15 Jun 2019. PMLR.

\bibitem{dan}
Mingsheng Long, Yue Cao, Jianmin Wang, and Michael~I. Jordan.
\newblock Learning transferable features with deep adaptation networks.
\newblock In {\em Proceedings of the 32Nd International Conference on
  International Conference on Machine Learning - Volume 37}, ICML'15, pages
  97--105. JMLR.org, 2015.

\bibitem{cdan}
Mingsheng Long, Zhangjie Cao, Jianmin Wang, and Michael~I. Jordan.
\newblock Conditional adversarial domain adaptation.
\newblock In {\em Proceedings of the 32Nd International Conference on Neural
  Information Processing Systems}, NIPS'18, pages 1647--1657, USA, 2018. Curran
  Associates Inc.

\bibitem{rtn}
Mingsheng Long, Han Zhu, Jianmin Wang, and Michael~I. Jordan.
\newblock Unsupervised domain adaptation with residual transfer networks.
\newblock In {\em Proceedings of the 30th International Conference on Neural
  Information Processing Systems}, NIPS'16, pages 136--144, USA, 2016. Curran
  Associates Inc.

\bibitem{jan}
Mingsheng Long, Han Zhu, Jianmin Wang, and Michael~I. Jordan.
\newblock Deep transfer learning with joint adaptation networks.
\newblock In {\em Proceedings of the 34th International Conference on Machine
  Learning - Volume 70}, ICML'17, pages 2208--2217. JMLR.org, 2017.

\bibitem{sntg}
Yucen Luo, Jun Zhu, Mengxi Li, Yong Ren, and Bo Zhang.
\newblock Smooth neighbors on teacher graphs for semi-supervised learning.
\newblock In {\em The IEEE Conference on Computer Vision and Pattern
  Recognition (CVPR)}, June 2018.

\bibitem{boostDAbyDLD}
M. {Mancini}, L. {Porzi}, S.~R. {Bulò}, B. {Caputo}, and E. {Ricci}.
\newblock Boosting domain adaptation by discovering latent domains.
\newblock In {\em 2018 IEEE/CVF Conference on Computer Vision and Pattern
  Recognition}, pages 3771--3780, June 2018.

\bibitem{mansour09}
Yishay Mansour, Mehryar Mohri, and Afshin Rostamizadeh.
\newblock Domain adaptation: Learning bounds and algorithms.
\newblock In {\em {COLT} 2009 - The 22nd Conference on Learning Theory,
  Montreal, Quebec, Canada, June 18-21, 2009}, 2009.

\bibitem{tpn}
Y. {Pan}, T. {Yao}, Y. {Li}, Y. {Wang}, C. {Ngo}, and T. {Mei}.
\newblock Transferrable prototypical networks for unsupervised domain
  adaptation.
\newblock In {\em 2019 IEEE/CVF Conference on Computer Vision and Pattern
  Recognition (CVPR)}, pages 2234--2242, June 2019.

\bibitem{mada}
Zhongyi Pei, Zhangjie Cao, Mingsheng Long, and Jianmin Wang.
\newblock Multi-adversarial domain adaptation.
\newblock In {\em Association for the Advancement of Artificial Intelligence
  (AAAI)}, pages 3934--3941, 2018.

\bibitem{SimNet}
P.~O. {Pinheiro}.
\newblock Unsupervised domain adaptation with similarity learning.
\newblock In {\em 2018 IEEE/CVF Conference on Computer Vision and Pattern
  Recognition}, pages 8004--8013, June 2018.

\bibitem{it_cluster_uda}
Ariya Rastrow, Frederick Jelinek, Abhinav Sethy, and Bhuvana Ramabhadran.
\newblock Unsupervised model adaptation using information-theoretic criterion.
\newblock In {\em Human Language Technologies: The 2010 Annual Conference of
  the North American Chapter of the Association for Computational Linguistics},
  HLT '10, pages 190--197, Stroudsburg, PA, USA, 2010. Association for
  Computational Linguistics.

\bibitem{dwt_mec}
S. {Roy}, A. {Siarohin}, E. {Sangineto}, S.~R. {Bulò}, N. {Sebe}, and E.
  {Ricci}.
\newblock Unsupervised domain adaptation using feature-whitening and consensus
  loss.
\newblock In {\em 2019 IEEE/CVF Conference on Computer Vision and Pattern
  Recognition (CVPR)}, pages 9463--9472, June 2019.

\bibitem{office31}
Kate Saenko, Brian Kulis, Mario Fritz, and Trevor Darrell.
\newblock Adapting visual category models to new domains.
\newblock In {\em Proceedings of the 11th European Conference on Computer
  Vision: Part IV}, ECCV'10, pages 213--226, Berlin, Heidelberg, 2010.
  Springer-Verlag.

\bibitem{adr}
Kuniaki Saito, Yoshitaka Ushiku, Tatsuya Harada, and Kate Saenko.
\newblock Adversarial dropout regularization.
\newblock In {\em International Conference on Learning Representations}, 2018.

\bibitem{mcd}
K. {Saito}, K. {Watanabe}, Y. {Ushiku}, and T. {Harada}.
\newblock Maximum classifier discrepancy for unsupervised domain adaptation.
\newblock In {\em 2018 IEEE/CVF Conference on Computer Vision and Pattern
  Recognition}, pages 3723--3732, June 2018.

\bibitem{gen_to_adapt}
S. {Sankaranarayanan}, Y. {Balaji}, C.~D. {Castillo}, and R. {Chellappa}.
\newblock Generate to adapt: Aligning domains using generative adversarial
  networks.
\newblock In {\em 2018 IEEE/CVF Conference on Computer Vision and Pattern
  Recognition}, pages 8503--8512, June 2018.

\bibitem{it_cluster_uda2}
Yuan Shi and Fei Sha.
\newblock Information-theoretical learning of discriminative clusters for
  unsupervised domain adaptation.
\newblock In {\em Proceedings of the 29th International Coference on
  International Conference on Machine Learning}, ICML'12, pages 1275--1282,
  USA, 2012. Omnipress.

\bibitem{dirt_t}
Rui Shu, Hung Bui, Hirokazu Narui, and Stefano Ermon.
\newblock A {DIRT}-t approach to unsupervised domain adaptation.
\newblock In {\em International Conference on Learning Representations}, 2018.

\bibitem{dada}
Hui Tang and Kui Jia.
\newblock Discriminative adversarial domain adaptation.
\newblock In {\em Association for the Advancement of Artificial Intelligence
  (AAAI)}, 2020.

\bibitem{adda}
E. {Tzeng}, J. {Hoffman}, K. {Saenko}, and T. {Darrell}.
\newblock Adversarial discriminative domain adaptation.
\newblock In {\em 2017 IEEE Conference on Computer Vision and Pattern
  Recognition (CVPR)}, pages 2962--2971, July 2017.

\bibitem{t_sne}
Laurens van~der Maaten and Geoffrey Hinton.
\newblock Visualizing data using t-sne.
\newblock {\em Journal of Machine Learning Research (JMLR)}, 9
  (Nov):2579–2605, 2008.

\bibitem{officehome}
Hemanth Venkateswara, Jose Eusebio, Shayok Chakraborty, and Sethuraman
  Panchanathan.
\newblock Deep hashing network for unsupervised domain adaptation.
\newblock In {\em 2017 IEEE Conference on Computer Vision and Pattern
  Recognition (CVPR)}, pages 5385--5394, 2017.

\bibitem{tada}
Ximei Wang, Liang Li, Weirui Ye, Mingsheng Long, and Jianmin Wang.
\newblock Transferable attention for domain adaptation.
\newblock In {\em Association for the Advancement of Artificial Intelligence
  (AAAI)}, 2019.

\bibitem{hla}
Jun Wen, Risheng Liu, Nenggan Zheng, Qian Zheng, Zhefeng Gong, and Junsong
  Yuan.
\newblock Exploiting local feature patterns for unsupervised domain adaptation.
\newblock In {\em Association for the Advancement of Artificial Intelligence
  (AAAI)}, 2019.

\bibitem{UnsupervisedEmbeddingICML16}
Junyuan Xie, Ross Girshick, and Ali Farhadi.
\newblock Unsupervised deep embedding for clustering analysis.
\newblock In {\em International Conference on Machine Learning - Volume 48},
  pages 478--487, 2016.

\bibitem{mstn}
Shaoan Xie, Zibin Zheng, Liang Chen, and Chuan Chen.
\newblock Learning semantic representations for unsupervised domain adaptation.
\newblock In Jennifer Dy and Andreas Krause, editors, {\em Proceedings of the
  35th International Conference on Machine Learning}, volume~80 of {\em
  Proceedings of Machine Learning Research}, pages 5423--5432,
  Stockholmsmässan, Stockholm Sweden, 10--15 Jul 2018. PMLR.

\bibitem{larger_norm}
R. {Xu}, G. {Li}, J. {Yang}, and L. {Lin}.
\newblock Larger norm more transferable: An adaptive feature norm approach for
  unsupervised domain adaptation.
\newblock In {\em 2019 IEEE/CVF International Conference on Computer Vision
  (ICCV)}, pages 1426--1435, Oct 2019.

\bibitem{GMVAE}
Linxiao Yang, Ngai-Man Cheung, Jiaying Li, and Jun Fang.
\newblock Deep clustering by gaussian mixture variational autoencoders with
  graph embedding.
\newblock In {\em The IEEE International Conference on Computer Vision (ICCV)},
  October 2019.

\bibitem{density_estimate}
Bianca Zadrozny.
\newblock Learning and evaluating classifiers under sample selection bias.
\newblock In {\em Proceedings of the Twenty-first International Conference on
  Machine Learning}, ICML '04, pages 114--, New York, NY, USA, 2004. ACM.

\bibitem{iCAN}
W. {Zhang}, W. {Ouyang}, W. {Li}, and D. {Xu}.
\newblock Collaborative and adversarial network for unsupervised domain
  adaptation.
\newblock In {\em 2018 IEEE/CVF Conference on Computer Vision and Pattern
  Recognition}, pages 3801--3809, June 2018.

\bibitem{symnets_v2}
Yabin Zhang, Bin Deng, Hui Tang, Lefei Zhang, and Kui Jia.
\newblock Unsupervised multi-class domain adaptation: Theory, algorithms, and
  practice.
\newblock {\em ArXiv}, abs/2002.08681, 2020.

\bibitem{partnet}
Y. {Zhang}, K. {Jia}, and Z. {Wang}.
\newblock Part-aware fine-grained object categorization using weakly supervised
  part detection network.
\newblock {\em IEEE Transactions on Multimedia}, pages 1--1, 2019.

\bibitem{mdd}
Yuchen Zhang, Tianle Liu, Mingsheng Long, and Michael Jordan.
\newblock Bridging theory and algorithm for domain adaptation.
\newblock In Kamalika Chaudhuri and Ruslan Salakhutdinov, editors, {\em
  Proceedings of the 36th International Conference on Machine Learning},
  volume~97 of {\em Proceedings of Machine Learning Research}, pages
  7404--7413, Long Beach, California, USA, 09--15 Jun 2019. PMLR.

\bibitem{metafgnet}
Yabin Zhang, Hui Tang, and Kui Jia.
\newblock Fine-grained visual categorization using meta-learning optimization
  with sample selection of auxiliary data.
\newblock In {\em The European Conference on Computer Vision (ECCV)}, September
  2018.

\bibitem{symnets}
Y. {Zhang}, H. {Tang}, K. {Jia}, and M. {Tan}.
\newblock Domain-symmetric networks for adversarial domain adaptation.
\newblock In {\em 2019 IEEE/CVF Conference on Computer Vision and Pattern
  Recognition (CVPR)}, pages 5026--5035, June 2019.

\bibitem{da_theory3}
Han Zhao, Remi Tachet~Des Combes, Kun Zhang, and Geoffrey Gordon.
\newblock On learning invariant representations for domain adaptation.
\newblock In Kamalika Chaudhuri and Ruslan Salakhutdinov, editors, {\em
  Proceedings of the 36th International Conference on Machine Learning},
  volume~97 of {\em Proceedings of Machine Learning Research}, pages
  7523--7532, Long Beach, California, USA, 09--15 Jun 2019. PMLR.

\end{thebibliography}
}

\clearpage
\appendix

%%%%%%%%% BODY TEXT
\section{Other implementation details}
\label{SecOtherImplement}
Other implementation details are as follows: 1) the momentum is set to $0.9$; 2) the weight decay is set to $0.0001$; 3) the batch size is set to $64$; 4) the number of training epochs is set to $200$; 5) for each trial, we follow \cite{cdan} and use the best-performing clustering model as the test model; 6) data augmentations of random crop and horizontal flip are applied during training; 7) the number of the task-specific FC layers of the base network is set to $2$ (i.e. $2048 \to 512 \to K$), where the first FC layer is the so-called bottleneck layer \cite{bsp, cdan, larger_norm}; 8) we perform discriminative clustering in the bottleneck feature space as additional regularization; 9) we implement our experiments in PyTorch. %; 9) we follow \cite{DeepClusterRelativeEMICCV17,UnsupervisedEmbeddingICML16} to initilize the introduced auxiliary distributions as $q_{i,k}^t={\rm I}[k=\hat{y}_i^t]$ and $\widetilde{q}_{i,k}^t={\rm I}[k=\hat{y}_i^t]$ for $i \in \{1, 2, \dots, n_t\}$, where $\hat{y}_i^t$ is the assigned label by standard $K$-means clustering on the instance features $\{\mathbf{z}_i^t\}_{i=1}^{n_t}$ extracted by the ImageNet pre-trained base network; 10) for the $K$-means, the target cluster centers are initialized as the class centroids of the source data; 11) at the first training epoch, the learnable cluster centers $\{\bm{\mu}_k\}_{k=1}^K$ are initialized based on the cluster assignments of $\{\mathbf{z}_i^t\}_{i=1}^{n_t}$ by the $K$-means (together with labeled source $\{\mathbf{z}_j^s\}_{j=1}^{n_s}$); 12) the $K$ target cluster centers for computing source weights by (12) in the main text are initialized by the $K$-means.

Our proposed SRDC simultaneously learns parameters of the feature embedding function $\mathbf{\theta}$, the classifier $\mathbf{\vartheta}$, and the learnable cluster centers $\{\mathbf{\mu}_k\}_{k=1}^K$ by minimizing the structurally regularized deep clustering objective (11). Note that we re-initialize $\{\mathbf{\mu}_k\}_{k=1}^K$ at the start of each training epoch based on the current cluster assignments of $\{\mathbf{z}_i^t\}_{i=1}^{n_t}$ together with labeled source $\{\mathbf{z}_j^s\}_{j=1}^{n_s}$; the introduced auxiliary distributions $q_{i,k}^t=\widetilde{q}_{i,k}^t={\rm I}[k=\hat{y}_i^t]$ for $i \in \{1, 2, \dots, n_t\}$ and $k \in \{1, 2, \dots, K\}$ at the first training epoch, where $\hat{y}_i^t$ is the assigned class label by standard $K$-means clustering on the embedded target features $\{\mathbf{z}_i^t\}_{i=1}^{n_t}$; the weights $\{w_j^s\}_{j=1}^{n_s}$ are set to $1$ at the first training epoch. For the $K$-means, the target cluster centers are initialized as the class centroids of the source data. Training algorithm of SRDC is given in Algorithm \ref{alg:srdc}. 

\begin{algorithm}[htb]
	\caption{Training algorithm for SRDC, $E$ denotes the training epoch, $I$ denotes the training iteration, $B_t$ and $B_s$ denote the mini-batches.}
	\label{alg:srdc}
	\begin{algorithmic}[1]
		\item[\textbf{Input:}] unlabeled target samples ${\cal{T}}=\{\mathbf{x}_i^t\}_{i=1}^{n_t}$; labeled source samples ${\cal{S}}=\{(\mathbf{x}_j^s, y_j^s)\}_{j=1}^{n_s}$
		\item[\textbf{Output:}] $\mathbf{\theta}, \mathbf{\vartheta}, \{\mathbf{\mu}_k\}_{k=1}^K$
		\State Initialize: $\mathbf{\theta}, \mathbf{\vartheta}, \{\mathbf{\mu}_k\}_{k=1}^K$, $q_{i,k}^t=\widetilde{q}_{i,k}^t={\rm I}[k=\hat{y}_i^t]$ for $i \in \{1, 2, \dots, n_t\}$ and $k \in \{1, 2, \dots, K\}$, $w_j^s=1$ for $j \in \{1, 2, \dots, n_s\}$, $E = 1$
		\While {not converge}
		\For{$I\gets 1, MAX\_ITER$} 
		\State Sample $B_t$ and $B_s$ from $\cal{T}$ and $\cal{S}$
		\If{E != 1}
		\State Compute $q_{ik}^t$ and $\widetilde{q}_{ik}^t$ by using (2)
		\EndIf
		\State Update $\mathbf{\theta}, \mathbf{\vartheta}, \{\mathbf{\mu}_k\}_{k=1}^K$ by minimizing (11) on $B_t$ and $B_s$
		\EndFor
		\State Compute $\{\mathbf{c}_k^t\}_{k=1}^K$ by standard $K$-means clustering
		\State Compute $w_j^s=1, j \in \{1, 2, \dots, n_s\}$ by using (12)
		\State Initialize: $\{\mathbf{\mu}_k\}_{k=1}^K$
		\State E = E + 1
		\EndWhile
	\end{algorithmic}
\end{algorithm}

\section{More comparisons}
\label{SecMoreResults}
\subsection{Comparisons on Office-31}
Comparisons with existing methods on Office-31 \cite{office31} using ResNet-50 \cite{resnet} as the base network are shown in Table \ref{table:results_office31}, where results of existing methods are quoted from their respective papers or the works of \cite{dsbn,tat,cdan,mada}. We can see that SRDC outperforms all compared methods on almost all transfer tasks, verifying the effectiveness of SRDC.

\subsection{Comparisons on ImageCLEF-DA}
Comparisons with existing methods on ImageCLEF-DA \cite{imageclefda} using ResNet-50 \cite{resnet} as the base network are reported in Table \ref{table:results_imageclefda}, where results of existing methods are quoted from their respective papers or the work of \cite{cdan,mada}. To compare our proposed SRDC with the state-of-the-art method CAN \cite{can} on ImageCLEF-DA, we report results of CAN obtained by running the official code (i.e. available at the website of  https://github.com/kgl-prml/Contrastive-Adaptation-Network-for-Unsupervised-Domain-Adaptation). We can see that SRDC exceeds all compared methods including CAN on all transfer tasks by a large margin, confirming the efficacy of SRDC.

\begin{table*}[!t]
	\begin{center}
		\begin{tabular}{|l|c|c|c|c|c|c|c|}
			\hline
			Method                & A $\rightarrow$ W & D $\rightarrow$ W & W $\rightarrow$ D & A $\rightarrow$ D & D $\rightarrow$ A & W $\rightarrow$ A & Avg \\
			
			\hline
			\hline
			Source Model \cite{resnet}                  & 77.8$\pm$0.2 & 96.9$\pm$0.1 & 99.3$\pm$0.1 & 82.1$\pm$0.2 & 64.5$\pm$0.2 & 66.1$\pm$0.2 & 81.1 \\
			
			RTN \cite{rtn}                   & 84.5$\pm$0.2 & 96.8$\pm$0.1 & 99.4$\pm$0.1 & 77.5$\pm$0.3 & 66.2$\pm$0.2 & 64.8$\pm$0.3 & 81.6 \\ %$\dagger$ 
			
			DAN \cite{dan}          & 81.3$\pm$0.3 & 97.2$\pm$0.0 & 99.8$\pm$0.0 & 83.1$\pm$0.2 & 66.3$\pm$0.0 & 66.3$\pm$0.1 & 82.3 \\
			
			DANN \cite{dann}       & 81.7$\pm$0.2 & 98.0$\pm$0.2 & 99.8$\pm$0.0 & 83.9$\pm$0.7 & 66.4$\pm$0.2 & 66.0$\pm$0.3 & 82.6 \\
			
			ADDA \cite{adda}                & 86.2$\pm$0.5 & 96.2$\pm$0.3 & 98.4$\pm$0.3 & 77.8$\pm$0.3 & 69.5$\pm$0.4 & 68.9$\pm$0.5 & 82.9 \\
			
			JAN-A \cite{jan}                   & 86.0$\pm$0.4 & 96.7$\pm$0.3 & 99.7$\pm$0.1 & 85.1$\pm$0.4 & 69.2$\pm$0.4 & 70.7$\pm$0.5 & 84.6 \\%$\dagger$ 
			
			MADA \cite{mada}                   & 90.0$\pm$0.1 & 97.4$\pm$0.1 & 99.6$\pm$0.1 & 87.8$\pm$0.2 & 70.3$\pm$0.3 & 66.4$\pm$0.3 & 85.2 \\%$\dagger$ 
			
			VADA \cite{dirt_t}              & 86.5$\pm$0.5 & 98.2$\pm$0.4 & 99.7$\pm$0.2 & 86.7$\pm$0.4 & 70.1$\pm$0.4 & 70.5$\pm$0.4 & 85.4 \\
			
			SimNet \cite{SimNet}               & 88.6$\pm$0.5 & 98.2$\pm$0.2 & 99.7$\pm$0.2 & 85.3$\pm$0.3 & 73.4$\pm$0.8 & 71.8$\pm$0.6 & 86.2 \\%$\star$ 
			
			GTA \cite{gen_to_adapt} & 89.5$\pm$0.5 & 97.9$\pm$0.3 & 99.8$\pm$0.4 & 87.7$\pm$0.5 & 72.8$\pm$0.3 & 71.4$\pm$0.4 & 86.5 \\
			
			MSTN \cite{mstn}                   & 91.3 & 98.9 & 100.0 & 90.4 & 72.7 & 65.6 & 86.5 \\%$\star$ from \cite{dsbn}
			
			MCD \cite{mcd}                  & 88.6$\pm$0.2 & 98.5$\pm$0.1 & \textbf{100.0}$\pm$0.0 & 92.2$\pm$0.2 & 69.5$\pm$0.1 & 69.7$\pm$0.3 & 86.5 \\
			
			SAFN+ENT \cite{larger_norm}   & 90.1$\pm$0.8 & 98.6$\pm$0.2 & 99.8$\pm$0.0 & 90.7$\pm$0.5 & 73.0$\pm$0.2 & 70.2$\pm$0.3 & 87.1 \\
			
			DAAA \cite{deepAdvAttentionAlign}  & 86.8$\pm$0.2 & 99.3$\pm$0.1 & 100.0$\pm$0.0 & 88.8$\pm$0.4 & 74.3$\pm$0.2 & 73.9$\pm$0.2 & 87.2 \\ %$\star$
			
			iCAN \cite{iCAN}                & 92.5 & 98.8 & \textbf{100.0} & 90.1 & 72.1 & 69.9 & 87.2 \\
			
			rRevGrad+CAT \cite{cat}   & 94.4$\pm$0.1 & 98.0$\pm$0.2 & 100.0$\pm$0.0 & 90.8$\pm$1.8 & 72.2$\pm$0.6 & 70.2$\pm$0.1 & 87.6 \\ %tensorflow
			
			CDAN+E \cite{cdan}  & 94.1$\pm$0.1 & 98.6$\pm$0.1 & \textbf{100.0}$\pm$0.0 & 92.9$\pm$0.2 & 71.0$\pm$0.3 & 69.3$\pm$0.3 & 87.7 \\
			
			MSTN+DSBN \cite{dsbn}              & 92.7 & 99.0 & 100.0 & 92.2 & 71.7 & 74.4 & 88.3 \\%$\star$ 
			
			TADA \cite{tada}                & 94.3$\pm$0.3 & 98.7$\pm$0.1 & 99.8$\pm$0.2 & 91.6$\pm$0.3 & 72.9$\pm$0.2 & 73.0$\pm$0.3 & 88.4 \\
			
			TAT \cite{tat}                  & 92.5$\pm$0.3 & \textbf{99.3}$\pm$0.1 & \textbf{100.0}$\pm$0.0 & 93.2$\pm$0.2 & 73.1$\pm$0.3 & 72.1$\pm$0.3 & 88.4 \\
			
			SymNets \cite{symnets}          & 90.8$\pm$0.1 & 98.8$\pm$0.3 & \textbf{100.0}$\pm$0.0 & 93.9$\pm$0.5 & 74.6$\pm$0.6 & 72.5$\pm$0.5 & 88.4 \\ %$\ddagger$ 
			
			BSP+CDAN \cite{bsp}             & 93.3$\pm$0.2 & 98.2$\pm$0.2 & \textbf{100.0}$\pm$0.0 & 93.0$\pm$0.2 & 73.6$\pm$0.3 & 72.6$\pm$0.3 & 88.5 \\
			
			MDD \cite{mdd}                  & 94.5$\pm$0.3 & 98.4$\pm$0.1 & \textbf{100.0}$\pm$0.0 & 93.5$\pm$0.2 & 74.6$\pm$0.3 & 72.2$\pm$0.1 & 88.9 \\%$\ddagger$
			
			DADA \cite{dada}                & 92.3$\pm$0.1 & 99.2$\pm$0.1 & \textbf{100.0}$\pm$0.0 & 93.9$\pm$0.2 & 74.4$\pm$0.1 & 74.2$\pm$0.1 & 89.0 \\
			
			CADA-P \cite{cada}              & \textbf{97.0}$\pm$0.2 & \textbf{99.3}$\pm$0.1 & \textbf{100.0}$\pm$0.0 & 95.6$\pm$0.1 & 71.5$\pm$0.2 & 73.1$\pm$0.3 & 89.5 \\ %$\star$  
			
			CAN \cite{can}                  & 94.5$\pm$0.3 & 99.1$\pm$0.2 & 99.8$\pm$0.2 & 95.0$\pm$0.3 & \textbf{78.0}$\pm$0.3 & 77.0$\pm$0.3 & 90.6 \\ %$\star$ 
			
			\hline
			\textbf{SRDC}                   & 95.7$\pm$0.2 & 99.2$\pm$0.1 & \textbf{100.0}$\pm$0.0 & \textbf{95.8}$\pm$0.2 & 76.7$\pm$0.3 & \textbf{77.1}$\pm$0.1 & \textbf{90.8} \\
			\hline
		\end{tabular}
	\end{center} 
	\caption{Results (\%) on Office-31 (ResNet-50). %Note that all compared methods are based on \textbf{PyTorch} implementation except for the methods in the last six rows based on \textbf{unknown} implementation, rRevGrad+CAT based on \textbf{TensorFlow} implementation, and RTN, JAN-A, and MADA based on \textbf{Caffe} implementation.
	}
	\label{table:results_office31} \vspace{1.0cm}
\end{table*}

\begin{table*}[ht]
	\begin{center}
		\resizebox{0.9\textwidth}{!}{
			\begin{tabular}{|l|c|c|c|c|c|c|c|}
				\hline
				Methods                 & I $\rightarrow$ P & P $\rightarrow$ I & I $\rightarrow$ C & C $\rightarrow$ I & C $\rightarrow$ P & P $\rightarrow$ C & Avg \\
				\hline
				\hline
				Source Model \cite{resnet}        & 74.8$\pm$0.3 & 83.9$\pm$0.1 & 91.5$\pm$0.3 & 78.0$\pm$0.2 & 65.5$\pm$0.3 & 91.2$\pm$0.3 & 80.7 \\
				
				DAN \cite{dan}                   & 74.5$\pm$0.4 & 82.2$\pm$0.2 & 92.8$\pm$0.2 & 86.3$\pm$0.4 & 69.2$\pm$0.4 & 89.8$\pm$0.4 & 82.5 \\
				
				RTN \cite{rtn}                   & 75.6$\pm$0.3 & 86.8$\pm$0.1 & 95.3$\pm$0.1 & 86.9$\pm$0.3 & 72.7$\pm$0.3 & 92.2$\pm$0.4 & 84.9 \\
				
				DANN \cite{dann}                 & 75.0$\pm$0.6 & 86.0$\pm$0.3 & 96.2$\pm$0.4 & 87.0$\pm$0.5 & 74.3$\pm$0.5 & 91.5$\pm$0.6 & 85.0 \\
				
				MADA \cite{mada}                 & 75.0$\pm$0.3 & 87.9$\pm$0.2 & 96.0$\pm$0.3 & 88.8$\pm$0.3 & 75.2$\pm$0.2 & 92.2$\pm$0.3 & 85.8 \\
				
				JAN \cite{jan}                   & 76.8$\pm$0.4 & 88.0$\pm$0.2 & 94.7$\pm$0.2 & 89.5$\pm$0.3 & 74.2$\pm$0.3 & 91.7$\pm$0.3 & 85.8 \\
				
				rRevGrad+CAT \cite{cat}          & 77.2$\pm$0.2 & 91.0$\pm$0.3 & 95.5$\pm$0.3 & 91.3$\pm$0.3 & 75.3$\pm$0.6 & 93.6$\pm$0.5 & 87.3 \\
				
				iCAN \cite{iCAN}                 & 79.5 & 89.7 & 94.7 & 89.9 & 78.5 & 92.0 & 87.4 \\
				
				CDAN+E \cite{cdan}               & 77.7$\pm$0.3 & 90.7$\pm$0.2 & 97.7$\pm$0.3 & 91.3$\pm$0.3 & 74.2$\pm$0.2 & 94.3$\pm$0.3 & 87.7 \\
				
				CAN \cite{can}                   & 77.2$\pm$0.6 & 90.3$\pm$0.5 & 96.0$\pm$0.2 & 90.9$\pm$0.3 & 78.0$\pm$0.6 & 95.6$\pm$0.6 & 88.0 \\
				
				CADA-P \cite{cada}               & 78.0 & 90.5 & 96.7 & 92.0 & 77.2 & 95.5 & 88.3 \\
				
				TAT \cite{tat}                   & 78.8$\pm$0.2 & 92.0$\pm$0.2 & 97.5$\pm$0.3 & 92.0$\pm$0.3 & 78.2$\pm$0.4 & 94.7$\pm$0.4 & 88.9 \\
				
				SAFN+ENT \cite{larger_norm}      & 79.3$\pm$0.1 & 93.3$\pm$0.4 & 96.3$\pm$0.4 & 91.7$\pm$0.0 & 77.6$\pm$0.1 & 95.3$\pm$0.1 & 88.9 \\
				
				SymNets \cite{symnets}           & 80.2$\pm$0.3 & 93.6$\pm$0.2 & 97.0$\pm$0.3 & 93.4$\pm$0.3 & 78.7$\pm$0.3 & 96.4$\pm$0.1 & 89.9 \\
				
				\hline
				\textbf{SRDC}       & \textbf{80.8}$\pm$0.3 & \textbf{94.7}$\pm$0.2 & \textbf{97.8}$\pm$0.2 & \textbf{94.1}$\pm$0.2 & \textbf{80.0}$\pm$0.3 & \textbf{97.7}$\pm$0.1 & \textbf{90.9} \\
				\hline
			\end{tabular}
		}
	\end{center} 
	\caption{Results (\%) on ImageCLEF-DA (ResNet-50). Note that results of CAN are obtained by running the official code.%Note that all compared methods are based on \textbf{PyTorch} implementation except for the method in the last row based on \textbf{unknown} implementation, rRevGrad+CAT based on \textbf{TensorFlow} implementation, and RTN and MADA based on \textbf{Caffe} implementation.
	}
	\label{table:results_imageclefda}
\end{table*}

\begin{table*}[!ht]
	\begin{center}
		\resizebox{1.0\textwidth}{!}{
			\begin{tabular}{|l|c|c|c|c|c|c|c|c|c|c|c|c|c|}%{p{4.5cm}p{0.5cm}<{\centering}p{0.5cm}<{\centering}p{0.5cm}<{\centering}p{0.5cm}<{\centering}p{0.5cm}<{\centering}p{0.5cm}<{\centering}p{0.5cm}<{\centering}p{0.5cm}<{\centering}p{0.5cm}<{\centering}p{0.5cm}<{\centering}p{0.5cm}<{\centering}p{0.5cm}<{\centering}p{0.5cm}<{\centering}}%
				\hline
				Methods                         & Ar$\rightarrow$Cl & Ar$\rightarrow$Pr & Ar$\rightarrow$Rw & Cl$\rightarrow$Ar & Cl$\rightarrow$Pr & Cl$\rightarrow$Rw & Pr$\rightarrow$Ar & Pr$\rightarrow$Cl & Pr$\rightarrow$Rw & Rw$\rightarrow$Ar & Rw$\rightarrow$Cl & Rw$\rightarrow$Pr    & Avg  \\
				%& Cl & Pr & Rw & Ar & Pr & Rw & Ar & Cl & Rw & Ar & Cl & Pr &  \\
				\hline
				\hline
				Source Model \cite{resnet}     & 34.9      & 50.0     & 58.0      & 37.4      & 41.9      & 46.2     & 38.5     & 31.2     & 60.4     & 53.9     & 41.2     & 59.9 & 46.1 \\
				
				DAN \cite{dan}                  & 43.6     & 57.0     & 67.9      & 45.8      & 56.5      & 60.4     & 44.0     & 43.6     & 67.7     & 63.1     & 51.5     & 74.3  & 56.3 \\
				
				DANN \cite{dann}     & 45.6     & 59.3     & 70.1      & 47.0      & 58.5      & 60.9     & 46.1     & 43.7     & 68.5     & 63.2     & 51.8      & 76.8 & 57.6 \\
				
				JAN \cite{jan}                  & 45.9     & 61.2     & 68.9      & 50.4      & 59.7      & 61.0     & 45.8     & 43.4     & 70.3     & 63.9     & 52.4      & 76.8 & 58.3 \\
				
				SE \cite{selfensembling}  & 48.8 & 61.8 & 72.8 & 54.1 & 63.2 & 65.1 & 50.6 & 49.2 & 72.3 & 66.1 & 55.9 & 78.7 & 61.5 \\%cited from {dwt_mec}
				
				DWT-MEC \cite{dwt_mec} & 50.3 & 72.1 & 77.0 & 59.6 & 69.3 & 70.2 & 58.3 & 48.1 & 77.3 & 69.3 & 53.6 & 82.0 & 65.6 \\%pytorch
				
				CDAN+E \cite{cdan}  & 50.7     & 70.6     & 76.0     & 57.6       & 70.0      & 70.0     & 57.4     & 50.9     & 77.3      & 70.9      & 56.7     & 81.6 & 65.8 \\
				
				TAT \cite{tat} & 51.6 & 69.5 & 75.4 & 59.4 & 69.5 & 68.6 & 59.5 & 50.5 & 76.8 & 70.9 & 56.6 & 81.6 & 65.8 \\
				
				BSP+CDAN \cite{bsp} & 52.0 & 68.6 & 76.1 & 58.0 & 70.3 & 70.2 & 58.6 & 50.2 & 77.6 & 72.2 & 59.3 & 81.9 & 66.3 \\
				
				SAFN \cite{larger_norm} & 52.0 & 71.7 & 76.3 & 64.2 & 69.9 & 71.9 & 63.7 & 51.4 & 77.1 & 70.9 & 57.1 & 81.5 & 67.3 \\
				
				TADA \cite{tada}       & 53.1 & 72.3 & 77.2 & 59.1 & 71.2 & 72.1 & 59.7 & 53.1 & 78.4 & 72.4 & 60.0 & 82.9 & 67.6 \\
				
				SymNets \cite{symnets} & 47.7 & 72.9 & 78.5 & 64.2 & 71.3 & 74.2 & 64.2 & 48.8 & 79.5 & 74.5 & 52.6 & 82.7 & 67.6 \\
				
				MDD \cite{mdd}         & 54.9 & 73.7 & 77.8 & 60.0 & 71.4 & 71.8 & 61.2 & 53.6 & 78.1 & 72.5 & 60.2 & 82.3 & 68.1 \\
				
				CAN \cite{can}         & \textbf{58.5} & 75.3 & 75.1 & 61.7 & 74.5 & 70.1 & 61.3 & \textbf{54.6} & 75.9 & 72.4 & 58.3 & 82.4 & 68.3 \\
				
				CADA-P \cite{cada}    & 56.9 & \textbf{76.4} & 80.7 & 61.3 & 75.2 & 75.2 & 63.2 & 54.5 & 80.7 & 73.9 & \textbf{61.5} & 84.1 & 70.2 \\%$\star$ 
				
				\hline
				\textbf{SRDC} & 52.3 & 76.3 & \textbf{81.0} & \textbf{69.5} & \textbf{76.2} & \textbf{78.0} & \textbf{68.7} & 53.8 & \textbf{81.7} & \textbf{76.3} & 57.1 & \textbf{85.0} & \textbf{71.3} \\
				\hline
			\end{tabular}
		}
	\end{center} 
	\caption{Results (\%) on Office-Home (ResNet-50). Note that results of CAN are obtained by running the official code.%Note that all compared methods are based on \textbf{PyTorch} implementation except for the method in the last row based on \textbf{unknown} implementation.
	}
	\label{table:results_officehome}
\end{table*}

\subsection{Comparisons on Office-Home}
Comparisons with existing methods on Office-Home using ResNet-50 \cite{resnet} as the base network are reported in Table \ref{table:results_officehome}, where results of existing methods are quoted from their respective papers or the works of \cite{cdan, dwt_mec}. To compare our proposed SRDC with the state-of-the-art method CAN \cite{can} on Office-Home, we report results of CAN obtained by running the official code (i.e. available at the website of  https://github.com/kgl-prml/Contrastive-Adaptation-Network-for-Unsupervised-Domain-Adaptation). We can observe that SRDC achieves much better results than all compared methods including CAN on almost all transfer tasks, affirming the usefulness of SRDC. 

\end{document}